%% file: M2F.tex
\documentclass{article}

\usepackage{multirow}

\usepackage{microtype}
\usepackage{graphicx}
\usepackage{subcaption}
\usepackage{booktabs}
\usepackage{tikz}
\usetikzlibrary{arrows.meta, positioning}
\usepackage{algorithm}
\usepackage[noend]{algorithmic}

\usepackage{amsmath}
\usepackage{amssymb}
\usepackage{mathtools}
\usepackage{amsthm}

\usepackage[most]{tcolorbox}
\usepackage{listings}
\usepackage{tableof}
\AtBeginDocument{\tofOpenTocFileForWrite}
\usepackage{float}
\definecolor{keywordcolor}{rgb}{0.7, 0.1, 0.1}
\definecolor{commentcolor}{rgb}{0.4, 0.4, 0.4}
\definecolor{stringcolor}{rgb}{0.5, 0.3, 0.2}
\definecolor{symbolcolor}{rgb}{0.1, 0.2, 0.7}
\definecolor{sortcolor}{rgb}{0.1, 0.5, 0.1}
\definecolor{attributecolor}{rgb}{0.7, 0.1, 0.1}
\definecolor{errorcolor}{rgb}{1, 0, 0}

\usepackage{hyperref}
\usepackage[capitalize,noabbrev]{cleveref}
\usepackage{enumitem}
\makeatletter
\renewcommand{\algorithmiccomment}[1]{\hspace{0.35em}{\scriptsize\textit{(#1)}}}
\renewcommand{\algorithmicrequire}{\textbf{Input:}}
\renewcommand{\algorithmicensure}{\textbf{Output:}}
\makeatother
\usepackage[margin=1in]{geometry}
\usepackage[numbers]{natbib}
\usepackage{etoc}

\usepackage{amsmath}
\usepackage{amssymb}
\usepackage{mathtools}
\usepackage{amsthm}
\usepackage{makecell}
\usepackage{array}
\newcolumntype{C}[1]{>{\centering\arraybackslash}p{#1}}

\usepackage{xspace}

\theoremstyle{plain}

\theoremstyle{definition}

\theoremstyle{remark}

\newcommand{\Lean}{\textsc{Lean}\xspace}

\newcommand{\PSR}{\textsc{PSR}\xspace}

\newcommand{\TableStyle}{\footnotesize\setlength{\tabcolsep}{3.5pt}\renewcommand{\arraystretch}{1.07}}

\newcommand{\paperkeywords}{autoformalization, theorem proving, verification, program repair, agents}
\hypersetup{pdfkeywords={\paperkeywords}}

\begin{document}

\title{M2F: Automated Formalization of Mathematical Literature at Scale}

\author{
Zichen Wang$^{1}$, Wanli Ma$^{2}$, Zhenyu Ming$^{3}$, Gong Zhang$^{3}$, Kun Yuan$^{4}$, Zaiwen Wen$^{2}$\thanks{Corresponding author} \\
\small $^{1}$School of Mathematical Sciences, Peking University \\
\small $^{2}$Beijing International Center for Mathematical Research, Peking University \\
\small $^{3}$Huawei Technologies\\
\small $^{4}$ Center for Machine Learning Research, Peking University\\
\small \texttt{zichenwang25@stu.pku.edu.cn}, \texttt{wlma@pku.edu.cn},\\
\small \texttt{mingzhenyu1@huawei.com}, \texttt{nicholas.zhang@huawei.com}\\
\small\texttt{kunyuan@pku.edu.cn}, \texttt{wenzw@pku.edu.cn}
}
\date{}
\maketitle

\begin{abstract}
Automated formalization of mathematics enables mechanical verification but remains limited to isolated theorems and short snippets. Scaling to textbooks and research papers is largely unaddressed, as it requires managing cross-file dependencies, resolving imports, and ensuring that entire projects compile end-to-end. We present M2F (Math-to-Formal), the first agentic framework for end-to-end, project-scale autoformalization in Lean. The framework operates in two stages. The statement compilation stage splits the document into atomic blocks, orders them via inferred dependencies, and repairs declaration skeletons until the project compiles, allowing placeholders in proofs.  The proof repair stage closes these holes under fixed signatures using goal-conditioned local edits. Throughout both stages, M2F keeps the verifier in the loop, committing edits only when toolchain feedback confirms improvement. 
In approximately three weeks, M2F converts long-form mathematical sources into a project-scale Lean library of 153,853 lines from 479 pages textbooks on real analysis and convex analysis, fully formalized as Lean declarations with accompanying proofs. This represents textbook-scale formalization at a pace that would typically require months or years of expert effort. On FATE-H, we achieve $96\%$ proof success (vs.\ $80\%$ for a strong baseline). Together, these results demonstrate that practical, large-scale automated formalization of mathematical literature is within reach. The full generated Lean code from our runs is available at \url{https://github.com/optsuite/ReasBook.git}.

\end{abstract}

\section{Introduction}
\label{sec:intro}
Most mathematical knowledge resides in textbooks and research papers. Formalizing even a single section of such material in \Lean \citep{moura2015lean,moura2021lean4} remains labor-intensive, typically demanding domain experts and, at scale, coordinated curation and annotation by teams. The bottleneck is project-level compilation, not any individual proof: imports and namespaces must be chosen consistently, definitions must typecheck under a pinned library snapshot, dependencies must be stabilized across files, and the code must be iteratively repaired until the project elaborates. This work dominates the cost of formalization and slows the growth of libraries such as \texttt{mathlib} \citep{mathlib2020} and \texttt{OptLib} \citep{optlib3,optlib1,optlib2}. Prior work provides key ingredients but does not directly address this \emph{project-scale compilation} problem. Neural theorem provers and agentic systems focus on closing goals \emph{given} well-formed statements within a buildable project; autoformalization translates informal text and LaTeX into formal snippets but struggles to maintain global buildability as documents grow and structural errors accumulate. We review these lines of work in \S\ref{sec:related}. Our focus is the upstream task: turning textbook- and paper-scale sources into a consistent Lean project that elaborates end-to-end.

\begin{figure*}[t]
  \centering
  \includegraphics[width=\textwidth]{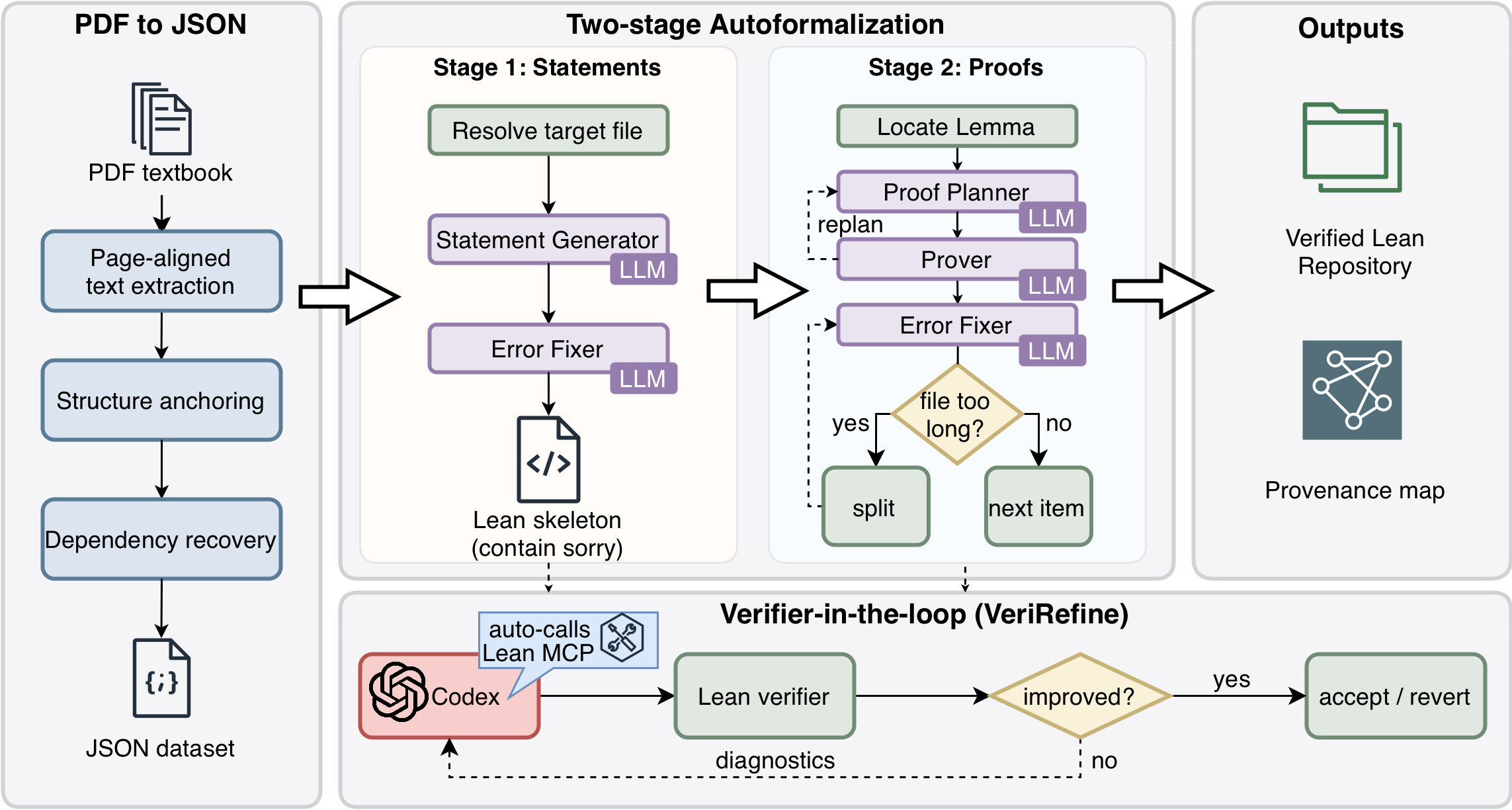}
  \caption{The M2F pipeline for project-scale automated formalization.}
  \label{fig:m2f_pipeline}
\end{figure*}

We address this gap with \textbf{M2F} (short for Math-to-Formal), a framework for automated formalization of mathematical literature at project scale. We formulate the task as \emph{knowledge compilation} under a fixed Lean environment $\mathcal{E}$: given a textbook or paper as input, produce a Lean project that elaborates end-to-end and provides span-level provenance linking declarations back to the source text. Our core methodology is \emph{verifier-certified refinement} (VeriRefine): M2F proposes localized edits and commits only when Lean’s toolchain feedback certifies strict progress on compilation, or—once compilation holds—on closing remaining proof holes. This accept-if-improves rule makes progress measurable and prevents regressions from accumulating over long runs. M2F instantiates VeriRefine in two stages. Stage~1 (statement compilation) decomposes documents into atomic blocks, infers and refines a dependency schedule, emits declaration skeletons, and repairs them until the project elaborates, permitting \texttt{sorry} placeholders in proofs. Stage~2 (proof repair) freezes statement signatures and closes remaining holes via goal-conditioned local edits. We evaluate Stage~2 under matched statements and matched compute using verifier-normalized budgets (verifier calls). Our experiments demonstrate both document-scale compilation and strong prover behavior. Across long-form sources spanning \emph{312 pages} of real analysis, \emph{140 pages} of convex analysis, and \emph{27 pages} of research-paper exposition, M2F produces end-to-end buildable Lean artifacts under $\mathcal{E}$ \emph{in a single end-to-end run}, reaching \emph{241 files}, \emph{4{,}116 declarations}, and \emph{153{,}853 lines of Lean} (Table~\ref{tab:proj_stats}); under matched statements, Stage~2 closes all proof holes on these compiled projects (PSR$=100\%$, proof success rate under matched statements). We additionally evaluate Stage~2 on FATE-H because it provides fully formal Lean statements and serves as a standard anchor for controlled prover comparison; replacing each target proof with \texttt{sorry} yields a clean matched-statement protocol that isolates proof repair. Under this protocol, Stage~2 achieves \emph{96\%} PSR on FATE-H, outperforming a state-of-the-art prover baseline (Seed-Prover~1.5 at \emph{80\%}, a \emph{+16-point} gain; Table~\ref{tab:fateh_psr}). Our contributions are summarized as follows.
\begin{itemize}
\item \textbf{Autoformalization at scale as knowledge compilation.}
We formulate textbook-/paper-scale formalization as compiling a buildable Lean project under a pinned environment, producing span-level provenance and enabling end-to-end evaluation beyond per-snippet correctness.

\item \textbf{VeriRefine.} We introduce an accept/revert refinement primitive governed solely by Lean toolchain feedback: edits are committed only when they reduce elaboration errors or, once elaboration succeeds, close proof holes. This ensures reliable progress under verifier-normalized compute budgets.
\item \textbf{M2F as a strong prover.}
We show that Stage~2 of M2F can serve as a strong prover under matched statements: using a CLI+MCP toolchain-in-the-loop with goal-conditioned local proof edits, it achieves state-of-the-art success on FATE-H among the compared provers, providing an external validation of the methodology.
\end{itemize}
Taken together, the three contributions above constitute M2F: a project-level knowledge-compilation formulation with provenance, a verifier-certified refinement primitive (VeriRefine), and a two-stage instantiation that scales to textbook- and paper-length sources while yielding strong prover behavior under matched statements. By keeping the Lean toolchain in the loop and using verifier feedback as the sole acceptance signal, M2F produces buildable Lean projects end-to-end and systematically reduces remaining proof holes without accumulating regressions, demonstrating that large-scale formalization is practically feasible under pinned environments. The M2F pipeline from end-to-end is shown in Figure~\ref{fig:m2f_pipeline}. 
More information on the M2F system is available in \url{https://github.com/optsuite/M2F.git}, and the Lean projects generated from our runs can be checked in \url{https://github.com/optsuite/ReasBook.git}. This positions proof assistants as reliable verifier infrastructure and suggests a path to accelerating the growth of formal mathematics libraries and downstream mathematical reasoning in AI systems. 

\section{Related Work}
\label{sec:related}

\textbf{Neural theorem proving in Lean.}
A large line of work studies automated proving when given goals that already elaborate in Lean, typically combining retrieval, search, and verifier feedback.
Representative tool interfaces and benchmarks include MiniF2F and LeanDojo \citep{zheng2022minif2f,yang2023leandojo}, and recent systems scale performance via stronger models and increasingly agentic search loops \citep{xin2024deepseekproverv15,ren2025deepseekproverv2,chen2025seedprover15,jiang2025fateformalbenchmarkseries,ospanov2025apollo}.
These approaches commonly assume a buildable project and well-typed statements under a pinned snapshot; in contrast, M2F targets an upstream regime where failures are dominated by project- and library-level structure (imports, name resolution, dependency ordering, and type-mismatch cascades) that arise before proof search is meaningful.

\textbf{Autoformalization with tool feedback and structural guidance.}
LLM-based autoformalization translates informal sources into formal statements and proofs \citep{wu2022autoformalization}, and Lean-centric pipelines increasingly incorporate compiler/tool feedback during generation and refinement \citep{lu2024process}.
Recent work further exploits structural cues and dependency information, including compiler-feedback refinement and consistency validation (ATF) \citep{guo2025atf}, DAG/dependency-aware proof workflows (ProofFlow) \citep{cabral2025proofflow}, retrieval with dependency-graph driven iterative autoformalization (Aria) \citep{wang2025ariaagentretrievaliterative}, and structure-to-instance theorem autoformalization (SITA) \citep{li2025sitaframeworkstructuretoinstancetheorem}.
M2F is complementary: we make end-to-end project buildability under $\mathcal{E}$ the primary objective and operationalize refinement through a verifier-certified accept/revert rule defined directly over toolchain feedback and remaining proof holes.

\textbf{Faithfulness, evaluation, and diagnostic-conditioned repair.}
Datasets and evaluation protocols for autoformalization include aligned corpora and alignment metrics \citep{gao2025herald,ying2024leanworkbook,azerbayev2023proofnet,gulati2024lean4autoformalbench,lu2024formalalign,poiroux2024reliableevalautoformalization}.
We complement these directions with a project-level evaluation protocol that tracks end-to-end buildability, uses provenance links to support faithfulness auditing beyond typechecking, and evaluates proof repair under matched statements while reporting compute in verifier-normalized units.
Methodologically, our accept/revert refinement loop is related to diagnostic-conditioned program repair, where candidate edits are proposed and validated by executing a checker and inspecting diagnostics \citep{monperrus2018repair,yasunaga2020repair,wang-etal-2024-intervenor,bouzenia2024repairagent,tang2024rex}; proof assistants add structure because ``compilation'' corresponds to type-theoretic elaboration under a pinned snapshot, and proof repair can further condition on elaborated goal states.
\section{Problem Setup and Pipeline Overview}
\label{sec:problem}

Figure~\ref{fig:m2f_pipeline} summarizes M2F as a project-scale, verifier-in-the-loop pipeline for automated formalization.
The input is a long-form LaTeX document (or a collection of textbook/paper sections) $\mathcal{D}$, and the output is a Lean project $\mathcal{P}$ that elaborates end-to-end under a pinned dependency revision.
M2F is organized as a two-stage pipeline (statements then proofs) executed inside a refinement protocol: the Lean toolchain returns range-annotated feedback, and each bounded edit is accepted only when verifier feedback certifies objective improvement.

\subsection{Input, Environment, and Output}

The input is a long-form source $\mathcal{D}$ (a LaTeX document or a collection of textbook/paper sections), normalized into an ordered sequence of JSON items by the PDF/LaTeX-to-JSON preprocessing in Figure~\ref{fig:m2f_pipeline} (page-aligned extraction, structure anchoring, and dependency recovery). This representation supports deterministic iteration over items and stable provenance linking. We use provenance spans to trace generated declarations back to the source; let $\mathrm{Span}(\mathcal{D})$ denote the set of character spans in $\mathcal{D}$. All verification is performed under a fixed Lean environment $\mathcal{E}$, consisting of a Lean toolchain version and a pinned dependency revision (e.g., a specific \texttt{mathlib} commit). Fixing $\mathcal{E}$ makes build outcomes and diagnostics comparable across runs; we assume determinism under $\mathcal{E}$, i.e., repeated verification produces identical diagnostics up to irrelevant ordering.

We expose the toolchain through project-level and file-level verification oracles:
$$
\textsc{VerifyProj}_{\mathcal{E}}(\mathcal{P}) \rightarrow (ok,\Delta), 
\textsc{VerifyFile}_{\mathcal{E}}(\mathcal{P},f) \rightarrow (ok,\Delta_f).
$$
Here $\Delta$ and $\Delta_f$ are finite multisets of diagnostics annotated with source ranges in the corresponding Lean project or file; each diagnostic includes a severity tag (error/warning/info) and a message string. We write $\mathrm{err}(\cdot)$ for the number of error-level diagnostics (defined for both $\Delta$ and $\Delta_f$), and treat $ok$ as shorthand for $\mathrm{err}(\Delta)=0$ (project) or $\mathrm{err}(\Delta_f)=0$ (file); warnings do not affect success. Note that provenance spans $\mathrm{Span}(\mathcal{D})$ live in the source coordinate system, while diagnostic ranges live in Lean file coordinates.

The output is a Lean project $\mathcal{P}$ together with a provenance map
$
\pi:\mathrm{Decl}(\mathcal{P}) \rightarrow 2^{\mathrm{Span}(\mathcal{D})},
$
where $\mathrm{Decl}(\mathcal{P})$ is the set of globally named declarations introduced across modules/files. Stage~1 permits \texttt{sorry} in proof bodies but requires that all modules/imports elaborate successfully (zero error-level diagnostics); Stage~2 optionally targets a placeholder-free project with no remaining \texttt{sorry}.

Although we expose $\textsc{VerifyProj}$ for reporting end-to-end buildability (e.g., \texttt{lake build}), the refinement loop itself is certified at file granularity: each attempted edit re-invokes $\textsc{VerifyFile}$ on the touched file and is accepted/reverted based solely on toolchain outputs. This separation makes the pipeline portable across agents and runtimes: patch proposal can use LLMs and tool queries, while certification depends only on $\mathcal{E}$ and verifier feedback.
\subsection{Two-Stage Pipeline}

M2F separates knowledge import from discharging proof obligations: at document scale, proof repair is meaningful only after imports, namespaces, typing, and dependencies have been stabilized at the project level. 

In Stage~1 (statements), M2F maps $\mathcal{D}$ to a project $\mathcal{P}_1$ that elaborates end-to-end under $\mathcal{E}$ while allowing proof placeholders. Each JSON item is assigned a target file; an LLM proposes a Lean declaration skeleton (possibly containing \texttt{sorry}); and when compilation fails the system proposes bounded patches guided by localized diagnostics. After each patch, it re-runs $\textsc{VerifyFile}$ and commits the edit only if verifier feedback certifies objective improvement; otherwise it reverts. 

In Stage~2 (proofs), M2F takes $\mathcal{P}_1$ and reduces remaining proof holes to obtain $\mathcal{P}_2$ under matched statements (statement signatures are held fixed; edits are restricted to proofs and optional local helpers that do not change existing signatures). Given the next proof item, the system locates the corresponding Lean declaration and its hole location (typically the inserted \texttt{sorry}) to establish the target file/range for repair. It then iterates a proof planner/prover to propose candidate proof patches; failed attempts trigger bounded repair via an error fixer and replanning. To keep verification stable at scale, the pipeline includes a deterministic guard for oversized files that triggers splitting before continuing. Both stages share the same refinement primitive (VeriRefine): propose a localized edit, re-run $\textsc{VerifyFile}$, and accept/revert based on verifier-certified objective improvement. We report compute in verifier-normalized units (the number of $\textsc{VerifyFile}$ calls), which reduces sensitivity to wall-clock variability and provides a uniform budget across long-running compilation and proof repair.
\section{Method: VeriRefine for M2F}
\label{sec:method}

VeriRefine is an operator for Lean projects under a fixed environment $\mathcal{E}$ (toolchain + pinned dependencies).
A state is a project $\mathcal{P}$.
Each refinement step proposes a bounded patch $p$ to a file region and commits it \emph{only if} Lean verification certifies a strict improvement of an explicit objective; otherwise the step is rolled back.
This accept/revert contract decouples \emph{how patches are proposed} (which may use language models, retrieval, and structured tool feedback) from \emph{how progress is certified} (which depends only on verifier outputs), making long-run refinement stable under verifier-call budgets.
\subsection{Oracles, Diagnostics, and Scopes}

Fix a Lean environment $\mathcal{E}$. For a project state $\mathcal{P}$ and file $f$, the file-level verifier returns
$$\textsc{VerifyFile}_{\mathcal{E}}(\mathcal{P},f)\rightarrow(ok,\Delta_f),$$
where $\Delta_f$ is a finite multiset of diagnostics and we treat
$ok=\textsf{true}$ as shorthand for $\mathrm{err}(\Delta_f)=0$ (warnings ignored), i.e., $f$ elaborates under $\mathcal{E}$.
We model each diagnostic as $d=(r,\kappa,\mu)$ with source range $r$ in $f$, severity $\kappa\in\{\textsf{error},\textsf{warning},\textsf{info}\}$, and message string $\mu$; writing $\mathrm{rng}(d)$, $\mathrm{kind}(d)$, and $\mathrm{msg}(d)$ for projections, the error count is
$\mathrm{err}(\Delta_f)=|\{d\in\Delta_f:\mathrm{kind}(d)=\textsf{error}\}|$.

A scope $S$ is a finite union of ranges in $f$, and diagnostic localization is
$$\textsc{Localize}(\Delta_f,S)=\{d\in\Delta_f:\mathrm{rng}(d)\cap S\neq\emptyset\}$$ with
$\mathrm{err}_S(\Delta_f,S)=\mathrm{err}(\textsc{Localize}(\Delta_f,S))$.
Let $\textsc{Hdr}(f)$ be a bounded header region (imports/namespace/section) and $\textsc{Decl}(c)$ the declaration range of a newly inserted skeleton $c$.
We use $\textsc{Scope}(c,f)=\textsc{Decl}(c)\cup\textsc{Hdr}(f)$ and, for a diagnostic or hole $x$, $\textsc{Scope}(x,f)=\mathrm{rng}(x)\cup\textsc{Hdr}(f)$.

When $\mathrm{err}(\Delta_f)=0$, we optionally query an elaborated goal state at a hole $h$ via
$$\textsc{GoalState}_{\mathcal{E}}(\mathcal{P},f,h)\in\{(g,\Gamma),\bot\},$$
returning goal type $g$ and local context $\Gamma=(x_1:T_1,\ldots,x_m:T_m)$, or $\bot$ if unavailable.

\subsection{Objectives and Verifier-Certified Accept/Revert}

We compare objective pairs by a priority order:
$(a,b)\prec(a',b')$ iff $a<a'$ or $(a=a'\wedge b<b')$.
For file diagnostics $\Delta_f$, define the primary metric $E(\Delta_f)=\mathrm{err}(\Delta_f)$.
Stage~1 uses the secondary metric $L(\Delta_f,S)=\mathrm{err}_S(\Delta_f,S)$ for the current scope $S$;
Stage~2 uses the secondary metric $H(f)=|\mathcal{H}(f)|$, where $\mathcal{H}(f)$ are occurrences of \texttt{sorry} outside comments/strings.
This ordering ensures that we never accept a patch that increases compilation errors, even if it would reduce locality errors or hole count. Target scheduling is external to acceptance. Stage~1 (resp.\ Stage~2) processes statement (resp.\ proof) items in increasing JSON index order, and each proof item may include a reference answer used only to condition patch proposals.

All edits are executed through $\textsc{TryPatch}_s$. Given $s\in\{1,2\}$, state $\mathcal{P}$, file $f$, scope $S$, a candidate patch $p$, and current diagnostics $\Delta_f$, it snapshots $f$, applies $p$ restricted to $S$, and re-runs $\textsc{VerifyFile}_{\mathcal{E}}(\mathcal{P},f)$ to obtain $\Delta_f'$.
It commits iff the stage objective strictly improves in the priority order:
Stage~1 accepts iff $(E(\Delta_f'),L(\Delta_f',S))\prec(E(\Delta_f),L(\Delta_f,S))$ and
Stage~2 accepts iff $(E(\Delta_f'),H'(f))\prec(E(\Delta_f),H(f))$, where $H'(f)$ is the hole count after applying $p$.
Otherwise it rolls back.
Because acceptance requires strict improvement, the sequence of accepted patches is monotone with respect to the stage objective, preventing oscillation and making verifier-call budgets a direct bound on attempted edits.

\subsection{Patch-Proposal Operators (Instantiation Layer)}
\label{sec:operators}

VeriRefine separates \emph{proposal} from \emph{certification}.
We describe proposal as a family of operators that may use tool feedback (diagnostics, goal states) and auxiliary context (e.g., reference proof text), while certification is always enforced by $\textsc{TryPatch}_s$.
Concretely, M2F uses the following operator types:

\begin{itemize}
\item \textsc{GenSkeleton}: $(u,\mathcal{P},f)\mapsto c$ inserts a declaration skeleton for a statement item $u$.
\item \textsc{RepairPatch}: $(\mathcal{P},f,\delta)\mapsto p$ proposes a localized patch conditioned on localized diagnostics $$\delta=\textsc{Localize}(\Delta,S).$$
\item \textsc{FixCompileError}: $(\mathcal{P},f,d)\mapsto p$ proposes a patch targeting a selected error diagnostic $d$ when $$\mathrm{err}(\Delta)>0.$$
\item \textsc{Plan}/\textsc{Replan}: $(u,g,\Gamma)\mapsto \pi$ proposes (and updates) a proof plan; $u$ may include a reference answer.
\item \textsc{ProposeProofPatch}: $(\mathcal{P},f,h,\pi,u,g,\Gamma)\mapsto p$ proposes a hole-closing proof patch within $\textsc{Scope}(h,f)$.
\item \textsc{SplitIfLargeAndResolve} (optional): given a long file, split at declaration boundaries while preserving namespace/section context and a linear import chain among parts, then re-resolve the target location by label.
\end{itemize}

\subsection{Preprocessing: Input as Ordered JSON Items}
\label{sec:preprocessing}

M2F takes as input an ordered stream of JSON items derived from long-form sources.
Statement items provide a label and a LaTeX span describing the statement/definition environment; proof items share the same index space and additionally carry an \texttt{proof} field containing the source proof text (LaTeX/NL).
In this paper we treat preprocessing as producing these ordered, provenance-carrying JSON items; the semantic schema and invariants of these items are specified in Appendix~\ref{app:json_schema}.

\subsection{Statement Compilation and Proof Repair}

Stage~1 compiles ordered statement items into a Lean project $\mathcal{P}_1$ that elaborates under $\mathcal{E}$ while allowing proof placeholders.
For each statement item, the system inserts a Lean skeleton into its target file and iteratively applies localized repairs until the file elaborates (or the per-item budget is exhausted); all candidate edits are mediated by $\textsc{TryPatch}_1$ and are committed only when verifier feedback certifies improvement.
When a required name and type must be made available before its body is stabilized (e.g., ordering constraints or latent cycles in long-form exposition), Stage~1 admits typed stubs that introduce the intended constant with its intended type while deferring the actual body, so that downstream items can elaborate.
Concretely, we use stub templates of the following form, where $n_b$ and $T_b$ denote the intended name and type for the current block:
\begin{tcolorbox}[
  colback=gray!10,
  colframe=black,
  boxrule=0.5pt,
  arc=1mm,
  left=2mm, right=2mm,
  top=0.2mm, bottom=0.2mm
]
\begin{lstlisting}
theorem n_b : T_b := by sorry
lemma   n_b : T_b := by sorry
def     n_b : T_b := sorry
abbrev  n_b : T_b := by sorry
example : T_b := by sorry
instance n_b : T_b := by sorry
\end{lstlisting}
\end{tcolorbox}Algorithm~\ref{alg:statement} summarizes the verifier-certified statement compilation procedure.

\begin{algorithm}[t]
\caption{Verifier-certified statement compilation}
\label{alg:statement}
{\footnotesize
\begin{algorithmic}
\REQUIRE Ordered statement items $\mathcal{I}^{\text{stmt}}$ (JSON); environment $\mathcal{E}$; per-item budget $K$
\ENSURE Compilable project $\mathcal{P}_1$; provenance map $\pi$
\STATE $\mathcal{P}\gets\textsc{InitProject}(\mathcal{E})$;\; $\pi\gets\emptyset$
\FOR{$u\in\mathcal{I}^{\text{stmt}}$ \textbf{in increasing index order}}
  \STATE $f\gets\textsc{TargetFile}(u)$;\; $snap_0\gets\textsc{Snapshot}(\mathcal{P},f)$
  \STATE $(c,\pi_u)\gets\textsc{GenSkeleton}(u,\mathcal{P},f)$;\\ $\textsc{Insert}(\mathcal{P},f,c)$;\; $\pi\gets\pi\cup\pi_u$
  \STATE $S\gets\textsc{Scope}(c,f)$;\; $(\_,\Delta)\gets\textsc{VerifyFile}_{\mathcal{E}}(\mathcal{P},f)$;\; $t\gets 0$
  \WHILE{$\mathrm{err}(\Delta)>0 \land t<K$}
    \STATE $\delta\gets\textsc{Localize}(\Delta,S)$
    \IF{$\delta=\emptyset$}
      \STATE $S\gets\textsc{ExpandScope}(S,\Delta)$
    \ELSE
      \STATE $p\gets\textsc{RepairPatch}(\mathcal{P},f,\delta)$
      \STATE $(\_,\Delta)\gets\textsc{TryPatch}_1(\mathcal{P},f,S,p,\Delta)$
    \ENDIF
    \STATE $t\gets t+1$
  \ENDWHILE
  \IF{$\mathrm{err}(\Delta)>0$} \STATE $\textsc{Restore}(\mathcal{P},f,snap_0)$ \ENDIF
\ENDFOR
\STATE \textbf{return} $\mathcal{P},\pi$
\end{algorithmic}}
\end{algorithm}

Stage~2 starts from the compilable project $\mathcal{P}_1$ and reduces the number of remaining proof holes while preserving elaboration.
We evaluate Stage~2 under matched statements: statement signatures are held fixed and only proofs (and optional local helpers that do not change existing signatures) may be edited.
Stage~2 processes proof items in increasing JSON index order; each item locates its target hole in the current Lean project (typically a labeled \texttt{sorry}) and may provide a reference answer (source proof text) used only to condition planning and patch proposal.
Proof repair alternates between compile-error recovery (if $\mathrm{err}(\Delta)>0$) and hole-focused proof patching with optional goal/context queries, with bounded retries and replanning under verifier-call budgets.
Algorithm~\ref{alg:proof} specifies the verifier-certified proof repair procedure for a single proof item.

\begin{algorithm}[t]
\caption{Verifier-certified proof repair for a proof item}
\label{alg:proof}
{\footnotesize
\begin{algorithmic}
\REQUIRE Project $\mathcal{P}_1$; proof item $u$ (label + optional answer); budgets $T$ (verifier calls), $R$ (retries), $C$ (replans)
\ENSURE Updated project state
\STATE $\mathcal{P}\gets\mathcal{P}_1$
\STATE $f\gets\textsc{TargetFile}(u)$;\\ $f\gets\textsc{SplitIfLargeAndResolve}(f,u)$
\STATE $(\_,\Delta)\gets\textsc{VerifyFile}_{\mathcal{E}}(\mathcal{P},f)$;\; $t\gets 0$
\WHILE{$t<T$}
  \IF{$\mathrm{err}(\Delta)>0$}
    \STATE $d\gets\textsc{SelectError}(\Delta)$;\; $S\gets\textsc{Scope}(d,f)$
    \STATE $p\gets\textsc{FixCompileError}(\mathcal{P},f,d)$
    \STATE $(\_,\Delta)\gets\textsc{TryPatch}_2(\mathcal{P},f,S,p,\Delta)$;\; $t\gets t+1$
    \STATE \textbf{continue}
  \ENDIF
  \STATE $h\gets\textsc{LocateTargetHole}(f,u)$
  \IF{$h=\bot$} \STATE \textbf{return} $\mathcal{P}$ \ENDIF
  \STATE $(g,\Gamma)\gets\textsc{GoalState}_{\mathcal{E}}(\mathcal{P},f,h)$ \COMMENT{optional}
  \STATE $\pi\gets\textsc{Plan}(u,g,\Gamma)$
  \FOR{$c=1$ \textbf{to} $C$}
    \FOR{$r=1$ \textbf{to} $R$}
      \STATE $q\gets\textsc{ProposeProofPatch}(\mathcal{P},f,h,\pi,u,g,\Gamma)$
      \STATE $(acc,\Delta)\gets\textsc{TryPatch}_2(\mathcal{P},f,\textsc{Scope}(h,f),q,\Delta)$;\; $t\gets t+1$
      \IF{$acc$} \STATE \textbf{return} $\mathcal{P}$ \ENDIF
      \IF{$t\ge T$} \STATE \textbf{return} $\mathcal{P}$ \ENDIF
    \ENDFOR
    \STATE $\pi\gets\textsc{Replan}(\pi,u,g,\Gamma,\Delta)$
  \ENDFOR
\ENDWHILE
\STATE \textbf{return} $\mathcal{P}$
\end{algorithmic}
}
\end{algorithm}

\section{M2F: Interface Contract}
\label{sec:arch}

M2F exposes document-scale autoformalization as a verifier-certified refinement loop (VeriRefine) over a pinned Lean environment $\mathcal{E}$ (Lean toolchain plus pinned dependency revision). Patch proposal can be implemented by any agent (LLM, retrieval, tool calls); certification is determined solely by toolchain outputs.

For a project state $\mathcal{P}$ and file $f$, the minimal oracle is file verification
$
\textsc{VerifyFile}_{\mathcal{E}}(\mathcal{P},f)\rightarrow(ok,\Delta_f),
$
returning a finite multiset of range-annotated diagnostics; we treat $ok$ as shorthand for $\mathrm{err}(\Delta_f)=0$ (warnings ignored). When $\mathrm{err}(\Delta_f)=0$, we optionally query an elaborated goal state at a hole $h$ via
$
\textsc{GoalState}_{\mathcal{E}}(\mathcal{P},f,h)\in\{(g,\Gamma),\bot\},
$
which is used only to condition patch proposals and never to decide acceptance. Edits are range-bounded. A patch is applied only within a scope $S$ (a finite union of ranges). We use the uniform constructor from \S\ref{sec:method}: for a newly inserted skeleton $c$, $S=\textsc{Decl}(c)\cup\textsc{Hdr}(f)$; for a diagnostic or hole $x$, $S=\mathrm{rng}(x)\cup\textsc{Hdr}(f)$. If $\textsc{Localize}(\Delta_f,S)=\emptyset$ but $\mathrm{err}(\Delta_f)>0$, we allow bounded scope expansion as a fallback. All edits are executed through a single commit/rollback wrapper: snapshot $\rightarrow$ apply within $S$ $\rightarrow$ re-verify $\rightarrow$ accept iff the stage objective strictly improves; otherwise rollback. Using the priority order from \S\ref{sec:method}, Stage~1 accepts iff
$(\mathrm{err}(\Delta_f'),\mathrm{err}_S(\Delta_f',S))\prec(\mathrm{err}(\Delta_f),\mathrm{err}_S(\Delta_f,S))$,
and Stage~2 accepts iff
$(\mathrm{err}(\Delta_f'),|\mathcal{H}'(f)|)\prec(\mathrm{err}(\Delta_f),|\mathcal{H}(f)|)$,
where $\mathcal{H}(f)$ counts remaining proof holes (e.g., \texttt{sorry}). Each attempted patch triggers exactly one call to $\textsc{VerifyFile}_{\mathcal{E}}$, so verifier-call budgets directly bound the number of attempted edits.
\section{Experiments}
\label{sec:exp}

We evaluate M2F on long-form mathematical sources and on the external benchmark FATE-H. On long-form sources, evaluation is end-to-end and fully checkable under a pinned Lean environment $\mathcal{E}$: M2F produces a Lean project that passes \texttt{lake build}, yielding a textbook-scale artifact of 241 files, 4{,}116 declarations, and 153{,}853 lines of Lean code from 479 pages of source material (Table~\ref{tab:proj_stats}). We then evaluate proof repair under matched statements on an audited statement layer whose signatures are manually verified against provenance-linked spans (Appendix~\ref{app:audit}); under this protocol, Stage~2 closes 875/875 audited proof holes while preserving elaboration (Table~\ref{tab:proj_stats} and Table~\ref{tab:stage2_text}). We additionally report results on FATE-H (100 problems), where statements already elaborate in Lean, so Stage~2 can be evaluated in isolation as a prover and compared by success rate.

\subsection{Setup}
\label{sec:exp:setup}

We evaluate on (a) textbooks: sections from \emph{Real Analysis} by Lebl~\citep{lebl2016real}\footnote{\scriptsize\url{https://github.com/jirilebl/real-analysis}.}
 and \emph{Convex Analysis} by Rockafellar~\citep{rockafellar1970convex}; (b) research papers: full sections with heterogeneous exposition and implicit assumptions; and (c) an external prover benchmark: FATE-H\footnote{\scriptsize\url{https://github.com/frenzymath/FATE-H.git}}.
We include FATE-H for three reasons.
First, its statements already elaborate in Lean, so Stage~1 is bypassed and Stage~2 can be evaluated under matched statements without confounding from statement generation.
Second, FATE-H is widely used in recent Lean prover evaluations, providing a standard anchor for success-rate comparisons when external systems do not expose comparable verifier-call accounting.
Third, it decouples evaluation from our document preprocessing and compilation stack, serving as a portability check that the same VeriRefine loop can be dropped into an existing Lean benchmark without corpus-specific heuristics.
Table~\ref{tab:datasets} summarizes scale; for long-form corpora, the Holes column counts all candidate proof locations after replacing proofs with \texttt{sorry}, and Stage~2 is evaluated on these matched-statement obligations after statement-signature audit.

\begin{table}[ht]
\caption{Corpora and scale. Blocks is the number of Stage~1 atomic blocks. Decls is the number of generated declarations after Stage~1. Holes is the number of candidate matched-statement proof obligations for Stage~2.}
\label{tab:datasets}
\centering
\TableStyle
\begin{tabular}{@{}lccccc@{}}
\toprule
Corpus & Sections & Pages & Blocks & Decls & Holes \\
\midrule
Real Analysis   & 36 & 312 & 416 & 1195 & 339 \\
Convex Analysis & 15 & 140 & 560 & 2620 & 499 \\
Paper           & 6 & 27  & 67  & 301  & 37  \\
\midrule
FATE-H          & -- & --  & --  & 100  & 100 \\
\bottomrule
\end{tabular}
\end{table}

All runs use the same Lean environment $\mathcal{E}$ (Lean: leanprover/lean4:v4.26.0-rc2) on the same hardware (Intel(R) Xeon(R) 6982P-C, 16 cores / 32 threads, 128 GiB RAM), and a file/project is considered verified iff Lean reports zero error-level diagnostics under $\mathcal{E}$.
LLM interactions are executed via Codex CLI with \texttt{model\_reasoning\_effort="high"}, and we interface with Lean through an LSP-based MCP server\footnote{\scriptsize\url{https://github.com/oOo0oOo/lean-lsp-mcp.git}} for lightweight tool queries during refinement (e.g., goal/context inspection when available); prompting philosophy and instrumentation are reported in Appendix~\ref{app:prompts_instr}.
A key implication for infrastructure is that our pipeline does not require large-scale human annotation to run: we do not assume human-labeled proof traces, dependency labels, or step-by-step demonstrations, and the primary signal is the Lean toolchain itself.
Finally, statement faithfulness is enforced by a provenance-linked manual audit that acts as an evaluation gate rather than a post-hoc sanity check: every theorem/lemma statement included in Stage~2 evaluation is compared by a domain expert against its provenance-linked source span, verifying preservation of quantifier structure, hypotheses (including implicit assumptions made explicit in prose), the conclusion, and essential typeclass/domain constraints; any statement that fails this checklist is excluded from the audited evaluation set (Appendix~\ref{app:audit}).
We do not use model-based statement checking as a correctness oracle because the statement layer is itself model-generated and model reviewers can share correlated failure modes (silently rubber-stamping subtle semantic drift); automated alignment is used only for triage, while human audit is the correctness criterion for the statement layer used in matched-statement proof repair.

\subsection{Project Artifacts and Statement Compilation}
\label{sec:exp:stage1}

The primary output of M2F is a buildable Lean library artifact rather than isolated proofs.
Table~\ref{tab:proj_stats} reports end-to-end artifacts after running the full pipeline on textbook- and paper-scale sources: for Lebl's \emph{Real Analysis} (312 pages) M2F produces 49 files, 1{,}195 declarations, and 34{,}327 lines of Lean code; for Rockafellar's \emph{Convex Analysis} it produces 164 files, 2{,}620 declarations, and 105{,}682 lines of Lean code.
Across all long-form corpora combined (479 pages), the output is 241 files and 153{,}853 lines of Lean code that build end-to-end under $\mathcal{E}$, indicating that imports, namespaces, and file/module structure have been stabilized at project scale without axiom declarations. 
The same table reports Stage2 progress under matched statements on the proof targets whose statement signatures pass the human audit. Table\ref{tab:datasets} reports corpus scale and the full set of candidate matched-statement proof obligations obtained by blanking proofs with \texttt{sorry}; in our current evaluation, all candidates pass audit, so the Hole counts coincide. For a qualitative view of artifact navigability and governance—workflow overview, ToC-faithful source-to-module alignment, and representative verified Lean excerpts—see Appendix~\ref{app:showcase}.

\begin{table*}[t]
\caption{End-to-end artifacts and proof progress under matched statements. PB indicates that the final project builds under $\mathcal{E}$. A concrete workflow walkthrough and code overview are provided in Appendix~\ref{app:showcase}.}
\label{tab:proj_stats}
\centering
\TableStyle
\begin{tabular}{@{}lcccccccc@{}}
\toprule
& \multicolumn{5}{c}{End-to-end artifact} & \multicolumn{3}{c}{Matched-statement Stage~2} \\
\cmidrule(lr){2-6}\cmidrule(lr){7-9}
Corpus & Blocks & PB & Files & Decls & LoC & Holes & Closed & PSR (\%) \\
\midrule
Real Analysis        & 416 & $\checkmark$ & 49  & 1195 & \textbf{34327}  & 339  & 339  & 100\\
Convex Analysis (Sec.~1--15) & 560 & $\checkmark$ & 164 & 2620 & \textbf{105682} & 499 & 499 & 100 \\
Paper                & 67  & $\checkmark$ & 28  & 301  & \textbf{13844}  & 37  & 37  & 100 \\
\midrule
Total (long-form) & 1043 & $\checkmark$ & 241 & 4116 & \textbf{153853} & 875 & 875 & 100 \\
\bottomrule
\end{tabular}
\end{table*}

We quantify Stage~1 using three metrics: SCC (statement compile coverage), the fraction of atomic blocks whose insertion and repair yields a file with zero error-level diagnostics; ARR (average repair rounds), the mean number of repair attempts per block (zero for blocks that elaborate immediately); and PB (project buildability). Compute is reported in verifier calls (one call is one invocation of file elaboration/typechecking via \texttt{lake env lean}); Stage~1 uses a per-block cap $K=3$ verify$\rightarrow$repair attempts.
Table~\ref{tab:stage1_main} shows SCC $=100\%$ and PB true on all long-form corpora, with ARR far below 1.0 (0.42/0.08/0.20), meaning that the refinement loop rarely needs to intervene beyond the generated skeletons; in particular, ARR $<0.5$ implies that strictly more than half of blocks require zero patch attempt, and the observed ARR values yield conservative lower bounds that at least 58\%/92\%/80\% of blocks elaborate immediately after insertion.
Overall, these results support a practical view of textbook-scale import as an infrastructure problem: most of the work is carried by toolchain-certified generation plus localized repair, while human effort is concentrated on lightweight review rather than dense manual authoring.

\begin{table}[ht]
  \caption{Stage~1 statement compilation. PB indicates whether the final project builds
  under $\mathcal{E}$ (allowing \texttt{sorry}).}
  \label{tab:stage1_main}
  \centering
  \TableStyle
  \begin{tabular}{@{}lccc@{}}
  \toprule
  Metric & Real Analysis & Convex Analysis & Paper \\
  \midrule
  SCC (\%)     & \textbf{100} & \textbf{100} & \textbf{100} \\
  ARR (rounds) & 0.42 & 0.08 & 0.20 \\
  PB           & $\checkmark$ & $\checkmark$ & \textbf{$\checkmark$} \\
  \bottomrule
  \end{tabular}
  \end{table}

\subsection{Proof Repair under Matched Statements}
\label{sec:exp:stage2}

Stage~2 evaluates proving capability isolated from statement generation under matched statements: starting from the same statement layer, each target proof is replaced by \texttt{sorry}, and methods may edit only proofs; PSR is the fraction of holes closed while preserving elaboration under $\mathcal{E}$.
To avoid conflating wall-clock variability with algorithmic behavior, we report compute in verifier calls; Stage~2 uses bounded retries and replanning ($R=10$, $C=21$), with totals reported in Appendix~\ref{app:cost}, and non-verifier oracle calls are not counted by default (Appendix~\ref{app:cost}).
The full Stage~2 closes all audited holes on all three long-form corpora (PSR $=100\%$; Table~\ref{tab:proj_stats}).
To isolate which components matter under a fixed statement layer, we run ablations on a representative slice of each long-form corpus:one fixed slice defined by a single dataset file per corpus from each of Real Analysis, Convex Analysis, and Paper, and PSR is computed on the audited holes within that slice.
Table~\ref{tab:stage2_text} reports the resulting PSR.
Because the sampled Real Analysis slice contains mostly routine obligations, several variants saturate at 100\% there; the slice is still informative for stress-testing the refinement loop itself, as one-shot repair drops sharply.
In contrast, the Paper slice is substantially more sensitive: diagnostics-only collapses (13.85\%), and disabling replanning also degrades performance, indicating that structured goal/context conditioning and plan revision are critical beyond local diagnostic repair when exposition is heterogeneous and dependencies are implicit.
\begin{table}[H]
\caption{Stage~2 ablations under matched statements on fixed slices (one dataset file per long-form corpus, selected and then fixed; slice identifiers are reported in Appendix~\ref{app:ablation_slices}). PSR is computed on holes within each slice; full-corpus results are in Table~\ref{tab:proj_stats}.}
\label{tab:stage2_text}
\centering
\TableStyle
\begin{tabular}{@{}lccc@{}}
\toprule
Method & Real Analysis  & Convex Analysis & Paper \\
\midrule
One-shot                 & 45.45 & 63.94 & 46.15 \\
Diagnostics-only         & 100   & 95.45 & 13.85 \\
No replanning ($C{=}1$)  & 100   & 90.91 & 46.97 \\
\textbf{M2F (full Stage~2)} & \textbf{100} & \textbf{100} & \textbf{100} \\
\bottomrule
\end{tabular}
\end{table}

On FATE-H, Stage~2 is evaluated in isolation (statements already elaborate): we replace each target proof with \texttt{sorry} and run proof repair under matched statements. Because external systems do not expose comparable verifier-call accounting, we compare PSR rather than verifier-normalized efficiency. Table~\ref{tab:fateh_psr} shows that M2F reaches 96\% fully automatically, improving by 16 points over the strongest reported Seed-Prover variant (80\%). We further define a reproducible light-supervision condition as an additional input artifact: a JSON lemma map with 31 entries, each consisting of (i) a fully-qualified Lean declaration name available under $\mathcal{E}$ and (ii) a one-sentence natural-language role description; no proof scripts, tactics, or intermediate derivations are provided. With the Stage~2 algorithm, budgets, and prompts unchanged (the lemma map is used only as extra conditioning context), the same pipeline solves one additional instance, reaching 97\%.

\begin{center}
\noindent
\begin{minipage}[t]{0.54\linewidth}
\vspace{0pt}
\centering
\includegraphics[width=\linewidth]{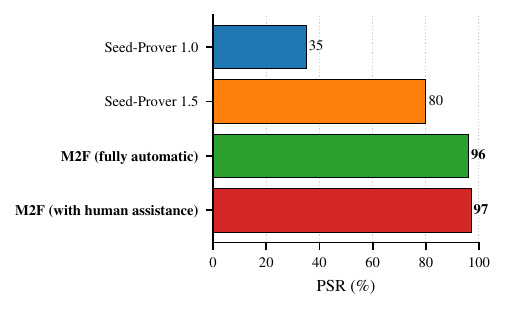}
\captionsetup{hypcap=false}
\captionof{figure}{PSR on FATE-H across provers.}
\label{fig:fateh_bar}
\end{minipage}\hfill
\begin{minipage}[t]{0.43\linewidth}
\vspace{0pt}
\centering
\captionsetup{hypcap=false}
\captionof{table}{Stage~2 on FATE-H (100 problems) under matched statements. Entries are PSR (\%). ``+31 decl lemma map'' is a reproducible light-supervision condition: a JSON lemma map with 31 fully-qualified Lean declaration names, each annotated by a one-sentence natural-language role description; no proof scripts or step-by-step traces are provided.}
\label{tab:fateh_psr}
\TableStyle
\begin{tabular}{@{}lc@{}}
\toprule
Method & FATE-H \\
\midrule
Seed-Prover~1.0 (medium)              & 35 \\
Seed-Prover~1.5 (agentic prover only) & 57 \\
Seed-Prover~1.5                       & 80 \\
\textbf{M2F (fully automatic)}        & \textbf{96} \\
\textbf{M2F (+31 decl lemma map)}     & \textbf{97} \\
\bottomrule
\end{tabular}
\end{minipage}
\end{center}
\subsection{Failure Analysis}
\label{sec:exp:failure}

On FATE-H, the four instances not solved automatically decompose cleanly by cause: one is solved under the reproducible +31-decl lemma-map condition; one is a benchmark issue whose formal Lean statement is inconsistent and therefore unprovable under $\mathcal{E}$; and two remain unsolved in our current runs without expert lemma maps. The benchmark issue was surfaced during M2F-assisted investigation: verifier-in-the-loop traces localized the inconsistency and enabled expert review to confirm the statement-level error, while the LLM-based semantic checks we tried (including GPT-5.2 Pro and other frontier models) did not flag the problem. This case illustrates how formal languages and proof-assistant verifiers complement large models: beyond proving, verifier-guided workflows can detect subtle statement errors and provide high-precision, mechanizable feedback signals that are useful for improving model reasoning and for curating reliable benchmarks.
\section{Conclusion}

We presented M2F, a verifier-in-the-loop pipeline that compiles textbook- and paper-scale sources into Lean projects that build end-to-end under a pinned environment while preserving span-level provenance. M2F combines dependency-aware statement compilation with matched-statement proof repair via goal-conditioned edits under a toolchain-certified accept/revert rule. On 479 pages, it produces a 153{,}853-LoC buildable library and closes all audited obligations; on FATE-H it solves 96\% fully automatically and 97\% with a 31-declaration natural-language lemma map. M2F also surfaced a FATE-H statement error missed by state-of-the-art LLM-based semantic checks (including GPT-5.2 Pro), underscoring the value of formal verification as both a prover oracle and a high-precision signal for evaluating and improving large-model mathematical reasoning. Overall, the remaining bottleneck shifts to natural-language grounding and library navigation.

\bibliography{main}
\bibliographystyle{icml2026}

\appendix

\onecolumn 

\input{appendix}
\end{document}

%% file: appendix.tex
\section{Prompting Philosophy and Instrumentation Details}
\label{app:prompts_instr}

This appendix documents two cross-cutting design choices that shape our system across all stages.
First, we summarize the prompting philosophy that we use to make the model produce \emph{canonical, mathlib-aligned} Lean statements and stable proofs.
Second, we describe the instrumentation semantics that make the pipeline auditable and enable the verifier-normalized compute accounting in Appendix~\ref{app:cost}.
We intentionally do not include literal prompt templates; instead we describe the design principles that guided prompt construction and the semantics of what is logged.

\subsection{Prompting philosophy}
\label{app:prompts_instr:philosophy}

Our prompting philosophy is organized around one principle: \textbf{statement-first formalization}.
Across stages, we treat ``writing the right statement'' as the most important driver of downstream success.
In practice, a statement that is minimal, canonical, and aligned with existing \texttt{mathlib} interfaces can dramatically reduce proof difficulty by avoiding accidental typeclass obligations, brittle goal shapes, and unnecessary rewriting overhead.

\noindent\textbf{Choosing the right declaration form.}
Prompts are designed to make the model choose standard Lean and \texttt{mathlib} declaration forms with canonical binders.
We encourage \texttt{def} for definitions of data/functions/predicates, \texttt{abbrev} when lightweight unfolding behavior is desirable, and \texttt{structure} for packaging fields when typeclass inference is not intended.
We use \texttt{class} only when downstream code should consume the structure via typeclass inference; otherwise \texttt{structure} is preferred.
For propositions, we use \texttt{theorem} for results explicitly presented as theorems in the source and default to \texttt{lemma} otherwise.
Finally, we reserve \texttt{instance} for canonical typeclass instances that should be found by inference, and we avoid inventing instances solely to make automation or rewriting easier.

\noindent\textbf{Avoid inline proofs inside terms.}
A stability rule in our prompts is to avoid embedding proofs inside larger terms.
Concretely, prompts discourage (and in some stages forbid) writing proof snippets such as \texttt{(by ...)} or \texttt{$\langle \dots, by \dots \rangle$} inside a term-level construction.
When a construction requires proof obligations (for example, closure properties for subtypes or structure fields), we design prompts to first introduce a separate helper \texttt{lemma} with a clean, minimal statement and then reference that lemma by name in the larger term.
This yields flatter goal structure, improves readability, and makes later repair and proof search more modular.

\noindent\textbf{Additional constraints for definitions.}
Definitions are handled more strictly than propositions.
Prompts are designed so that the model do not use tactic-mode proofs inside a \texttt{def} body; definitions should be expressed as short term constructions.
If a definition appears to require substantial proof work, prompts steer the model toward splitting: define the object/predicate with a clean \texttt{def} or \texttt{abbrev}, and move properties or obligations into separate lemmas.
When a definition must be left incomplete, prompts prefer \texttt{:= sorry} (not \texttt{by sorry}) to avoid introducing tactic blocks into definitions.

\noindent\textbf{Placeholders in Stage~1.}
Stage~1 allows placeholders to obtain a compilable statement layer.
Prompts enforce a consistent policy: proposition-valued declarations use \texttt{:= sorry} as a proof placeholder, and definitions use \texttt{:= sorry} if a placeholder is necessary.
We avoid using \texttt{axiom} to bypass obligations unless the global project rules explicitly allow it.

\noindent\textbf{A checklist for ``good statements''.}
Prompts implicitly encode (and our human review follows) a checklist that encourages canonical statements.
We reuse existing \texttt{mathlib} structures and predicates whenever possible instead of introducing ad-hoc ones, minimize binders and keep them canonical, separate object definitions from properties, avoid inline proofs inside terms by factoring helpers, and avoid non-canonical instances.

\noindent\textbf{Docstrings are required.}
Prompts require a short natural-language docstring immediately above each new declaration (\texttt{/-- ... -/} in Lean).
Docstrings are intended to improve maintainability and future reuse: they should briefly describe what the declaration is, why it exists, and (when applicable) how it relates to the source material.
We avoid copying long verbatim text from the source; concise descriptions are preferred.

\noindent\textbf{Proof stability and timeout robustness.}
When writing proofs (especially in Stage~2), prompts prioritize stable proof scripts and discourage forms that frequently trigger deterministic slowdowns.
We avoid long, deeply nested chains of \texttt{have} statements and prefer flatter structures using standard combinators such as \texttt{intro}, \texttt{refine}, \texttt{calc}, \texttt{rw}, and \texttt{simp}.
When a proof repeatedly requires many intermediate facts, prompts treat this as a signal that the statement or helper-lemma shape may be suboptimal; the model is encouraged to revise the statement layer or factor additional helpers rather than continuing a brittle proof search.

\noindent\textbf{Stage-specific intent.}
Stage~1 aims to produce a Lean project that elaborates under the pinned environment while allowing proof placeholders, so prompts emphasize canonical signatures, minimal assumptions, consistent placeholder policy, and diagnostics-driven repairs with small, local edits.
Stage~2 aims to eliminate target proof holes while preserving the audited statement layer, so prompts emphasize respecting matched statements (no signature edits), goal-directed proof search when goal/context is available, stable proof structure, and factoring nontrivial obligations into helper lemmas rather than embedding proofs inside terms.

\noindent\textbf{How we package context to the model.}
A central tradeoff is over-conditioning (too much context causing unnecessary refactors) versus under-conditioning (too little context causing missing assumptions or wrong types).
We adopt a minimal sufficient context principle: each prompt includes only the information needed for the current local decision.
Across stages, prompt inputs may include the focal content (an atomic block or a small neighborhood around a target hole), explicit edit-scope constraints, localized verifier diagnostics or goal/context information, and a compact shared header/import/notation context.
Windowing and truncation rules are fixed to keep prompts within model limits and to ensure repeatability; the corresponding artifact schemas that store prompt inputs are described in Appendix~\ref{app:json_schema}.

\subsection{Instrumentation semantics}
\label{app:instrumentation}

Instrumentation is designed so that reported metrics can be reconstructed from machine-readable logs rather than manual bookkeeping.
We log at attempt granularity: each model invocation and each verifier check produces a structured record, and runs also emit an append-only metrics stream.
This section describes \emph{what} we log and \emph{how} it is used; Appendix~\ref{app:json_schema} gives the corresponding JSON/JSONL schemas.

\subsubsection{Artifact types and their semantics}
\label{app:instrumentation:artifacts}

We distinguish three levels of artifacts.
First, per-call logs provide a full audit trail for each model call (inputs/outputs/diffs/tokens) and are primarily used for debugging and token backfill.
Second, compact JSON/JSONL stores provide structured, queryable traces: progress checkpoints for resumability, history stores for ``memory'' and debugging, and a metrics event log for analytics.
Third, optional rollups provide human-friendly summaries of persistent failures.

Table~\ref{tab:app:artifact_overview} summarizes the semantic role of each artifact category.
We emphasize semantic intent rather than on-disk paths; Appendix~\ref{app:json_schema} specifies the corresponding machine-readable formats.

\begin{table*}[t]
\caption{Instrumentation artifacts: content and primary use.}
\label{tab:app:artifact_overview}
\centering
\small
\setlength{\tabcolsep}{5pt}
\renewcommand{\arraystretch}{1.12}
\begin{tabular}{@{}>{\raggedright\arraybackslash}p{0.17\linewidth}>{\raggedright\arraybackslash}p{0.46\linewidth}>{\raggedright\arraybackslash}p{0.31\linewidth}@{}}
\toprule
Artifact & Contents & Primary use \\
\midrule
Per-call model log &
Full stdout/stderr for one model invocation, including effective prompt, tool-wrapper outputs, and emitted diffs; may also include token markers &
Per-call auditing/debugging; token backfill from stable markers \\

Progress checkpoint (JSON) &
A single resume cursor (e.g., next item or next file) plus minimal run identifiers &
Crash-safe resumability without reprocessing completed targets \\

History store (JSONL) &
Compact per-target records across retries (plan summaries, key errors, selected fields) &
Debugging and optional continuity context for later attempts \\

Metrics event log (JSONL) &
Append-only event stream with run/item/file boundaries, verifier checks, and oracle/model-call outcomes &
Primary source for paper metrics and verifier/oracle call accounting \\

Metrics summary (JSON) &
Run-end aggregates that mirror the final \texttt{run\_end} event payload &
Quick run-level accounting without scanning JSONL logs \\

Token backfill (JSONL) &
Per-target token totals reconstructed from stable token markers in per-call logs &
Robust token accounting when metrics events omit per-call token fields \\

Failure rollup (optional) &
Append-only human-readable list of persistent item/file failures with pointers &
Fast triage and rerun planning \\
\bottomrule
\end{tabular}
\end{table*}

\subsubsection{Recoverability: checkpoints and resumable runs}
\label{app:instrumentation:recover}

Long runs can be interrupted by timeouts, infrastructure instability, or budget limits.
To make experiments recoverable, each pipeline maintains a single-key JSON checkpoint that stores the next target to process.
We treat the exact key name and schema as part of the JSON artifact specification and describe it in Appendix~\ref{app:json_schema:checkpoint}; Appendix~\ref{app:instrumentation:recover} explains the operational semantics (when the cursor advances, when it stops, and how runs resume deterministically).

\noindent\textbf{Why checkpointing matters for accounting.}
Checkpointing guarantees that totals computed from logs can be reconstructed without ambiguity.
When a run is resumed, the resumed segment emits a new run id in the metrics stream, so corpus-level totals are computed by summing over the run ids included in the reported experiment.
This design prevents double counting and makes it clear which targets were attempted in each segment.

\subsubsection{Per-call logs: auditability and token extraction}
\label{app:instrumentation:percall}

Per-call logs are produced for every model invocation.
They are human-readable and capture the full transcript needed to reproduce what the model saw and did, including any diffs applied to the Lean project.

\noindent\textbf{Per-call log delimiters and token footer.}
Per-call logs begin with a literal \texttt{STDOUT:} section and then a literal \texttt{STDERR:} section.
Token usage is recorded as a two-line footer containing the literal marker \texttt{tokens used} followed by a decimal integer (commas may appear).
Our tooling parses this with a robust regex equivalent to \verb|tokens used\s*([0-9][0-9,]*)|.

\noindent\textbf{Audit scope.}
Because the per-call log contains the effective prompt (as printed by the Codex wrapper) and the corresponding stdout/stderr transcript, it provides a complete audit trail for debugging and for reconstructing token usage even when metrics events omit per-call token fields.
Per-call logs are also used to diagnose failures that involve toolchain behavior (e.g., unexpected verifier output or wrapper exceptions) that may not be captured in compact JSON summaries.

\subsubsection{Structured stores: history and metrics}
\label{app:instrumentation:structured}

\noindent\textbf{History store (JSONL).}
The history store is an append-only JSONL stream of compact per-task records.
Each record includes a timestamp, a run id, a task identifier, a kind tag (plan, replan request, fix attempt, warning cleanup, etc.), and a payload object.
The payload is designed to be lightweight and may truncate long strings to keep the store compact.
When history augmentation is enabled, a small window of recent records for the current file/task is loaded and included in later prompts to improve continuity across retries without requiring the full per-call logs.

\noindent\textbf{Metrics event log (JSONL).}
The metrics event log is the primary machine-readable instrumentation for paper metrics and accounting.
It is an append-only JSONL stream where each line is a JSON object with a timestamp, a run id, an event label, and an event payload.
Event labels cover run start/end, item/file start/end, verifier checks, and model-call results.
When present, event payloads may additionally include file snapshot summaries (size, modification time, line counts) that allow change-size analyses without reading full diffs.

\noindent\textbf{MCP/LSP tool queries are logged but not counted.}
During statement, proof, and final editing, the wrapper may issue Lean LSP/MCP queries (for example: diagnostics refreshes, goal inspection, hover/type information, local search, completions, and running small Lean snippets) to support interactive refinement.
These tool queries are recorded only in the raw per-call transcripts and are not emitted as structured metrics events.
As a result, the oracle-call totals in Appendix~\ref{app:cost}---which are derived from model-invocation events---do not include the frequency of LSP/MCP queries, and token totals do not include any cost attributable to these local tool queries.
In practice, LSP/MCP usage can contribute to wall-clock time and local compute, but it is excluded from our verifier-normalized and token-based accounting.

\subsubsection{Implementation notes: resumability and file-local verification}
\label{app:instrumentation:impl_notes}

All pipelines iterate deterministically over their target inventory (dataset indices for item-level stages, or a fixed file list for file-level cleanup).
To support interruption and restart, each pipeline maintains a single JSON checkpoint that stores the next target cursor; on restart, the pipeline reloads this checkpoint and resumes without re-processing completed targets.
This design avoids manual bookkeeping and enables reliable aggregation across runs: if a corpus/stage setting is completed across multiple resumptions, the reported totals simply sum across the corresponding run ids.

Verification is performed \emph{file-locally} using a pinned Lean environment: the pipeline repeatedly runs a single-file compile check (e.g., \texttt{lake env lean <file>}) on the currently edited file rather than rebuilding the entire project after every local edit.
This is both a performance choice and an accounting choice: each such single-file compile check is exactly one verifier call in our cost model (Appendix~\ref{app:cost:verifier_unit}), and the corresponding diagnostics provide the accept/reject signal for commit/rollback in all stages.

For very large generated sections, an optional splitter can refactor a file into multiple parts connected by a deterministic import chain (e.g., \texttt{sectionYY\_partK.lean} files imported by a thin aggregate module).
When splitting is enabled, proof repair targets only the containing part file, keeping both verification and patch scope local and reducing instability caused by overly large files.

\subsubsection{Mapping metrics events to paper metrics}
\label{app:instrumentation:metric_mapping}

The paper reports (i) success rates, (ii) verifier-normalized compute, (iii) oracle-call frequency, and (iv) token usage (as a complementary cost proxy).
All of these can be derived from the metrics event log and (for tokens) optionally corroborated via token backfill.

\noindent\textbf{Success and completion.}
For item-level pipelines, completion status is defined by item-end events that record whether the target item was processed successfully under budget.
For file-level final cleanup, completion status is defined by file-end and task-end events that record whether a file verified and how many \texttt{sorry} occurrences were eliminated.

\noindent\textbf{Verifier calls.}
Verifier-call totals are computed by counting verifier-check events (labeled \texttt{lean\_check} in our logs) and aggregating across the run ids included in a reported corpus/stage setting.
For some Stage~1 runs produced with an older metrics schema, verifier-check events may not be emitted explicitly; in that case we reconstruct verifier-call totals from per-item retry counters as described in Appendix~\ref{app:cost:total_calls}.
Appendix~\ref{app:cost} defines verifier calls as our primary compute unit and describes the aggregation procedure.

\noindent\textbf{Oracle calls and token usage.}
Oracle-call totals are computed by counting model-result events (agent result events) and summing across run ids, again matching the aggregation used for verifier calls.
For some Stage~1 runs produced with an older metrics schema, explicit repair-call events may not be emitted; in that case we reconstruct oracle-call totals from the same per-item retry counters used for verifier-call reconstruction (Appendix~\ref{app:cost:oracle}).
When per-call token counts are present in metrics, token totals can be computed by summing those fields.
To make accounting robust across logging variants, we also provide a token-backfill procedure that parses stable token markers from per-call logs and emits structured per-task token totals; Appendix~\ref{app:json_schema:token_backfill} specifies this artifact and Appendix~\ref{app:cost} explains how it is used for reporting.

\subsection{Human statement-audit protocol (faithfulness gate)}
\label{app:audit}

Stage~2 evaluation relies on a statement layer whose \emph{faithfulness} is checked against the original source.
For each long-form corpus (our textbook corpora and the paper corpus), theorem/lemma statements used in Stage~2 evaluation are manually audited by a human expert against the provenance-linked source excerpt shown during statement compilation.
This audit is the only correctness gate for statement faithfulness: model-based reviewers may be used for triage, but they are not treated as a guarantee because the statement layer is itself model-generated and reviewer models can share correlated failure modes.

\noindent\textbf{Audit checklist.}
Auditors compare the Lean declaration (binder structure, hypotheses, conclusion) against the corresponding source excerpt under standard informal conventions.
In particular, they check (i) quantifier structure and binder order, including implicit ``for all'' in prose; (ii) side conditions and domain restrictions (e.g., non-emptiness, measurability, integrability, boundedness); (iii) the logical strength of the conclusion; and (iv) whether any additional Lean-side assumptions were silently introduced (e.g., extra typeclass constraints) or omitted.
A statement is accepted iff the Lean signature is judged equivalent in mathematical content to the source excerpt.

\noindent\textbf{Coverage and corrections.}
For the textbook corpora used in our matched-statement Stage~2 evaluation, we manually audit every theorem/lemma statement that serves as a proof target.
In the current revision, this audit did not require statement-level edits: after checking the generated Lean signatures against the source excerpts, we found them faithful and therefore froze the matched statement layer as generated.
Because early runs did not maintain a separate per-statement audit ledger, we do not report a separate audited-statement count here; instead, the evaluation target counts (numbers of proof targets/holes) are reported in the main text (Table~\ref{tab:proj_stats}) and are the same targets used for the accounting totals in Appendix~\ref{app:cost}.

\noindent\textbf{Audit provenance and identifiers.}
Each audited statement is identified by the tuple \texttt{(data\_file, index, label)}: \texttt{data\_file} is recorded at run start, and \texttt{index} and texttt{label} are recorded at item start/end.
The source excerpt used for audit is exactly the dataset record payload (the \texttt{content} field together with any accompanying section \texttt{context} and \texttt{dependencies} when present) that was shown to the model during statement compilation; this payload is serialized into prompt metadata and preserved verbatim in the per-call model logs.
The audited Lean declaration is linked back to the same item via its \texttt{index} and texttt{label}, which also appears in the docstring/comment immediately preceding the generated declaration in the Lean file.
In our current datasets and logs, the excerpt is stored as a LaTeX string.

\subsection{Stage~2 ablation slices (section identifiers)}
\label{app:ablation_slices}

For Table~\ref{tab:stage2_text}, each long-form corpus uses a fixed ablation slice defined by a single dataset file (recorded as \texttt{data\_file}).
All ablation methods for that corpus are run on the same fixed slice.
We report slice identifiers by \texttt{data\_file} together with the observed \texttt{number\_components} coverage (where \texttt{number\_components = [chapter, section, \dots]}).

\begin{itemize}[leftmargin=*, itemsep=2pt, topsep=2pt, parsep=0pt, partopsep=0pt]
  \item \textbf{Real Analysis (Lebl).}
  \path{data/proof/ch-real-nums.json};
  observed coverage: chapter $=1$, section $\in \{1,2,3,5\}$.

  \item \textbf{Convex Analysis (Rockafellar).}
  \path{data/section04.json};
  observed coverage: chapter $=4$, section $\in \{1,2,3,4,5,6,7,8\}$.
  Note that despite the filename \path{section04.json}, this slice spans multiple sections within Chapter~4.

  \item \textbf{Paper.}
  \path{data/nesterov05_full.json};
  observed coverage: chapter $=1$, section $\in \{1,2,3,4,5\}$.
\end{itemize}

\noindent\textbf{Paper-only micro-ablation.}
We additionally ran a paper-only micro-ablation restricted to Chapter~1, Section~5 with slice id
\path{filtered__nesterov05_full__chap01_sec05.json}
(observed coverage: chapter $=1$, section $=5$).

\section{Semantic-First JSON Schemas for Intermediate Artifacts}
\label{app:json_schema}

This appendix specifies the machine-readable intermediate artifacts used for preprocessing inputs, resumability, debugging, and metrics reconstruction.
Appendix~\ref{app:instrumentation} describes the semantic role of these artifacts; here we focus on stable schema structure and invariants.
We intentionally avoid committing to any on-disk directory layout or filename convention.
Instead, we define each artifact by its semantic category (dataset record, checkpoint, history record, metrics event record, metrics summary, token backfill).

\subsection{Dataset record: a natural-language item}
\label{app:json_schema:dataset}

A dataset record is the canonical \emph{task input} describing one extracted mathematical unit (definition/lemma/theorem/etc.) in natural language (LaTeX-like).
The statement pipeline consumes the statement text to generate a Lean declaration skeleton.
The proof pipeline consumes the proof text (when present) to attempt proof completion for items that are meaningful proof targets (e.g., skipping definition-only records).

\noindent\textbf{Container format and ordering.}
A dataset is stored as a JSON array; each element is a record.
Consumers iterate deterministically by sorting records on \texttt{index}, even if the input is already sorted.
This makes partial runs resumable and ensures that logs are comparable across machines.

\noindent\textbf{Record schema (semantic view).}
Each record combines (i) stable identity, (ii) source position metadata, and (iii) the extracted content used for prompting.
Table~\ref{tab:app:dataset_keys} summarizes the keys and their meaning.

\begin{table}[ht]
\caption{Dataset record keys (semantic schema).}
\label{tab:app:dataset_keys}
\centering
\small
\setlength{\tabcolsep}{4pt}
\renewcommand{\arraystretch}{1.08}
\begin{tabular}{@{}p{0.29\linewidth}p{0.66\linewidth}@{}}
\toprule
Key & Meaning / used for \\
\midrule
\texttt{index} & Stable integer identifier; defines iteration order and is used as the primary resume cursor. \\
\texttt{label} & Human-readable label (e.g., ``Lemma 1.1''); used in logs and for locating targets. \\
\texttt{env} & Coarse type tag (e.g., definition vs.\ proposition); used for filtering which items are proof targets. \\
\texttt{number\_components} & Numbering decomposition used as a fallback for deterministic target positioning. \\
\texttt{extracted\_labels} & Source-level labels attached to the item; helpful for traceability and context. \\
\texttt{context} & Structured section metadata (chapter/section titles and indices) used for deterministic file/position inference. \\
\texttt{content} & Extracted statement text; primary input to statement compilation prompts. \\
\texttt{dependencies} & Lightweight dependency labels used as optional local context (consumer-specific). \\
\texttt{proof} & Extracted proof text; may be empty. Proof runs that require proof content skip empty-proof items. \\
\bottomrule
\end{tabular}
\end{table}

\noindent\textbf{Section metadata (\normalfont\texttt{context}).}
The \texttt{context} object records where the item appears in the source. In our runs it includes chapter/section titles together with numeric indices (chapter/section/subsection).
Values may be empty strings when the source lacks a corresponding heading.
Consumers use this metadata for deterministic target positioning and, in some configurations, for mapping items to a file/namespace layout.

\noindent\textbf{Invariants.}
The dataset file is a JSON array, \texttt{index} is unique, and consumers treat \texttt{index} order as the canonical iteration order.
If \texttt{proof} is empty, proof-stage consumers may skip the record depending on configuration (e.g., when proof text is required as reference context).

\noindent\textbf{Example record (redacted).}
\begin{tcolorbox}[
  colback=gray!10, colframe=black, boxrule=0.5pt, arc=1mm,
  left=2mm, right=2mm, top=0.4mm, bottom=0.4mm
]
\begin{lstlisting}[language={},basicstyle=\ttfamily\small]
{ "index": 1,
  "label": "Definition 1.1.1",
  "env": "def",
  "number_components": [1, 1, 1],
  "extracted_labels": ["def:1.1", "eq:1.1"],
  "context": {
    "chapter_number": 1,
    "chapter": "___",
    "section_number": "1",
    "section": "___",
    "subsection_number": "",
    "subsection": ""
  },
  "content": "\\begin{definition} ... \\end{definition}",
  "dependencies": [],
  "proof": "" }
\end{lstlisting}
\end{tcolorbox}

\subsection{Lemma-map supervision artifact (declaration-level hints)}
\label{app:json_schema:lemmap}

In the lightly supervised FATE-H setting, we use an optional \emph{lemma map} to provide minimal expert guidance for library navigation.
A lemma map consists only of declaration-level hints (names or short natural-language descriptions) and explicitly excludes any formal proof scripts, intermediate proof steps, or tactic traces.
It conditions planning/retrieval, but it does not affect acceptance: every edit is still accepted or rejected solely by verifier feedback under $\mathcal{E}$.

\noindent\textbf{Schema (semantic view).}
A lemma map can be represented as a JSON object keyed by problem id (or dataset index).
Each entry stores a list of relevant Lean declarations and an optional short note explaining why they are relevant.
The exact key names are not important; the semantics are.

\begin{tcolorbox}[
  colback=gray!10, colframe=black, boxrule=0.5pt, arc=1mm,
  left=2mm, right=2mm, top=0.4mm, bottom=0.4mm
]
\begin{lstlisting}[language={},basicstyle=\ttfamily\small]
{
  "problem_id": "FATEH_XX",
  "decl_hints": [
    "Mathlib.Analysis....",
    "Mathlib.Topology...."
  ],
  "notes": "optional natural-language rationale"
}
\end{lstlisting}
\end{tcolorbox}

\noindent\textbf{Consumption invariant.}
Downstream prompts are instructed to treat these hints as \emph{navigation cues} (which existing theorems/definitions to look at), not as proof steps.
In particular, the system may import, unfold, or apply the hinted declarations, but it must still produce a verifier-certified proof without relying on any unverified intermediate claims.

\subsection{Progress checkpoint: resumable cursor}
\label{app:json_schema:checkpoint}

A progress checkpoint stores the resume cursor for a pipeline.
It is a single JSON object with exactly one integer key.
Item-level pipelines store the next dataset index to process; file-level pipelines store the next file index to process.
Checkpoints are overwritten in-place and are intended to support resuming after interruption.

\noindent\textbf{Schema.}
Item-indexed pipelines use a key of the form \texttt{next\_index}.
File-indexed pipelines use a key of the form \texttt{next\_file\_index}.
The value is always the next target to process.

\subsection{History record: compact per-task trace}
\label{app:json_schema:history}

A history store is an append-only JSONL stream of compact records used for debugging and (optionally) prompt augmentation.
Each line is a single JSON object.
As described in Appendix~\ref{app:instrumentation}, the purpose of history is to provide lightweight ``memory'' without reading full per-call logs; here we specify the stable top-level schema and the most common record kinds.

\noindent\textbf{Top-level schema.}
Each history record includes a timestamp, a pipeline tag, a run id (for cross-reference into the metrics log), a Lean file identifier, a task identifier, a kind tag, an optional short summary, an optional log pointer, and a kind-specific payload object. Table~\ref{tab:app:history_top} summarizes the keys and their meaning.

\begin{table*}[ht]
\caption{History record fields (top-level schema).}
\label{tab:app:history_top}
\centering
\small
\setlength{\tabcolsep}{4pt}
\renewcommand{\arraystretch}{1.05}
\begin{tabular}{@{}p{0.22\linewidth}p{0.74\linewidth}@{}}
\toprule
Field & Meaning \\
\midrule
\texttt{ts} & UTC timestamp (ISO8601). \\
\texttt{pipeline} & Pipeline tag (e.g., proof or final). \\
\texttt{run\_id} & Run identifier linking to the metrics event log. \\
\texttt{lean\_file} & Lean file identifier (relative path or logical name). \\
\texttt{task\_id} & Task identifier (item index, file-index+line target, compile/warnings task, etc.). \\
\texttt{kind} & Kind tag indicating payload interpretation (plan, fix, replan request, warning cleanup, etc.). \\
\texttt{summary} & Optional short summary (may be truncated). \\
\texttt{log\_path} & Optional pointer to a per-call log for deep inspection. \\
\texttt{payload} & Kind-specific payload object (may truncate large strings). \\
\bottomrule
\end{tabular}
\end{table*}

\noindent\textbf{Common kinds.}
In our runs, common kinds include planner outputs, executor replan requests, repair attempts, and (in final cleanup) compilation-fix and warning-cleanup records.
Planner records typically store both a structured plan object (when parsable) and a raw plan text block; fix records typically store the motivating Lean error output (as a string) together with basic metadata such as attempt number and token usage.

\noindent\textbf{Truncation invariants.}
To keep history compact, the history writer may truncate long string fields (for example, summaries and large error logs).
Truncation is applied to top-level strings (including selected payload strings) and is intended as a space-saving mechanism; full raw text remains available in per-call logs when needed.

\noindent\textbf{Example history record (redacted).}
\begin{tcolorbox}[
  colback=gray!10, colframe=black, boxrule=0.5pt, arc=1mm,
  left=2mm, right=2mm, top=0.4mm, bottom=0.4mm
]
\begin{lstlisting}[language={},basicstyle=\ttfamily\small]
{ "ts": "2026-01-19T09:55:23.567889+00:00",
  "pipeline": "proof",
  "run_id": "proof_stage2_..._e70e21ee",
  "lean_file": "FormalPaper/.../section04_part1.lean",
  "task_id": "34",
  "kind": "agent_c_plan",
  "summary": "status=ok | main=smoothedObjective_lipschitz_gradient",
  "log_path": ".../proof_agent_c_...log",
  "payload": {
    "round": 1,
    "code": 0,
    "tokens_used": 12345,
    "model": "___",
    "reasoning_effort": "___",
    "plan": { "...": "..." },
    "plan_raw": "{...}"
  } }
\end{lstlisting}
\end{tcolorbox}

\subsection{Metrics event log: append-only run instrumentation}
\label{app:json_schema:metrics}

The metrics event log is the primary machine-readable instrumentation used for paper metrics and compute accounting.
It is an append-only JSONL stream where each line is a JSON object with four top-level fields: timestamp, run id, event label, and event payload.

\noindent\textbf{Top-level schema.}
Every event record has the same top-level structure:
\texttt{ts} (UTC timestamp), \texttt{run\_id} (identifier), \texttt{event} (event label), and \texttt{data} (payload object).
This invariant makes logs easy to parse and supports schema evolution by adding fields to \texttt{data} while preserving the top-level form.

\noindent\textbf{Event taxonomy.}
The precise set of \texttt{event} labels is implementation-defined but stable within a logging schema version.
For parsing and analysis, it is sufficient to group events into a small number of families: run boundaries, item/file boundaries, verifier checks, per-agent oracle-call results, and (for plan-driven runs) planning-loop bookkeeping events.
Appendix~\ref{app:instrumentation:metric_mapping} explains how these families are used to derive paper metrics; the key schema invariant here is that every line shares the same top-level structure and the \texttt{data} object contains enough identifiers to attribute the event to a target.

\noindent\textbf{Optional change-size snapshots.}
Some oracle-result events may include file snapshot summaries before and after applying an edit.
These snapshots record whether a file exists and, if it does, its size and a line-count summary.
They enable change-size analyses without parsing full diffs.

\noindent\textbf{Example metrics record.}
\begin{tcolorbox}[
  colback=gray!10, colframe=black, boxrule=0.5pt, arc=1mm,
  left=2mm, right=2mm, top=0.4mm, bottom=0.4mm
]
\begin{lstlisting}[language={},basicstyle=\ttfamily\small]
{ "ts": "2026-01-14T17:23:57.451963+00:00",
  "run_id": "proof_stage2_...",
  "event": "item_start",
  "data": { "index": 1, "label": "Lemma 1.1", "chapter": 1, "section": 1, "local_index": 1 } }
\end{lstlisting}
\end{tcolorbox}

\subsection{Metrics summary: end-of-run totals}
\label{app:json_schema:summary}

For convenience, each run writes a metrics summary as a single JSON object.
The summary mirrors the \texttt{run\_end} event payload and provides run-level totals without scanning the full JSONL stream.

\noindent\textbf{Common fields.}
All summaries include a pipeline identifier and a small set of run-level totals such as processed targets, resume cursor, total elapsed seconds, and aggregated attempt counters.
The exact set of counters depends on the pipeline, but the intent is consistent: the summary provides a concise accounting view for that run segment.

\noindent\textbf{Example summary (redacted).}
\begin{tcolorbox}[
  colback=gray!10, colframe=black, boxrule=0.5pt, arc=1mm,
  left=2mm, right=2mm, top=0.4mm, bottom=0.4mm
]
\begin{lstlisting}[language={},basicstyle=\ttfamily\small]
{ "pipeline": "final",
  "processed_files": 2,
  "next_file_index": 2,
  "total_seconds": 951.1927,
  "total_b_attempts": 1,
  "total_c_plans": 1,
  "total_a_attempts": 1,
  "total_sorries_eliminated": 1,
  "total_tokens_used": 123456 }
\end{lstlisting}
\end{tcolorbox}

\subsection{Token backfill: per-task token totals from per-call logs}
\label{app:json_schema:token_backfill}

Token usage can be reconstructed either from metrics events (when per-call token fields are present) or by parsing stable token markers from per-call logs.
To make accounting robust to logging variants, we provide a token-backfill procedure that parses per-call logs, aggregates token totals per task, and emits these totals as a dedicated metrics run.

\noindent\textbf{Backfill event schema.}
Backfill emits metrics records with event label \texttt{task\_tokens}.
Each record includes a stage tag, a task identifier, total tokens for that task, and a per-agent breakdown when multiple agents contribute to the same task.
The record may also include optional convenience fields such as the human label and Lean file identifier when available.

\noindent\textbf{Example backfill record (redacted).}
\begin{tcolorbox}[
  colback=gray!10, colframe=black, boxrule=0.5pt, arc=1mm,
  left=2mm, right=2mm, top=0.4mm, bottom=0.4mm
]
\begin{lstlisting}[language={},basicstyle=\ttfamily\small]
{ "event": "task_tokens",
  "data": {
    "stage": "final",
    "task": "0_L119",
    "tokens_used_total": 65831,
    "tokens_used_by_agent": { "a": 34170, "c": 31661 },
    "log_file_count": 2,
    "lean_file": "FormalBook/.../section01.lean" } }
\end{lstlisting}
\end{tcolorbox}

\section{Verifier-Normalized Compute and Cost Accounting}
\label{app:cost}

This appendix defines the compute and cost quantities reported in our experiments and explains how they are reconstructed from the structured artifacts described in Appendix~\ref{app:json_schema}.
Our primary compute unit is the number of Lean single-file compilation checks (\emph{verifier calls}), because all stages repeatedly apply candidate edits and then re-check the affected Lean file via a uniform pinned environment.
In addition, we report the frequency and token usage of model invocations (``oracle calls'') and include a conservative robustness check that charges each oracle call as a fixed fraction of a verifier call.

\subsection{Compute unit: verifier calls}
\label{app:cost:verifier_unit}

We measure compute in terms of \emph{verifier calls}.
A verifier call is one invocation of Lean elaboration/typechecking for a single file under the pinned environment $\mathcal{E}$.
Operationally, this corresponds to running a single-file compile check (e.g., \texttt{lake env lean <file>}) and collecting the resulting diagnostics.

\noindent\textbf{What is counted.}
We count \emph{every} verifier invocation performed by the method, including verifier calls inside repair/search loops and those triggered by any optional stabilization procedures.
Concretely, whenever a stage applies a candidate patch and then re-checks the file to decide accept/reject (or to advance to the next target), that re-check contributes one verifier call.

\noindent\textbf{Success criterion for accounting.}
A file is considered \emph{verified} iff Lean reports zero \emph{error-level} diagnostics under $\mathcal{E}$.
Warnings are ignored for verification accounting unless they are explicitly promoted to errors by the toolchain or by a pipeline configuration.

\noindent\textbf{Why this unit matches the pipeline design.}
All stages rely on repeated single-file compilation checks as the ground-truth accept/reject signal.
In the statement stage, the system appends a skeleton and compiles the target file; if compilation fails, it applies a small repair and compiles again.
In the proof and final stages, the system alternates between proposing proof edits and compiling to confirm that the file still elaborates (and that the targeted \texttt{sorry} is removed when required).
Because these checks are performed in a uniform pinned environment, verifier-call counts are comparable across corpora and stages.

\subsection{Budgets and per-target caps}
\label{app:cost:budgets}

Our pipelines are budgeted by placing explicit caps on the number of propose$\rightarrow$verify loops and on the number of fallback repairs, so that total compute is predictable and runs are resumable.
This subsection restates the caps used in our experiments and fixes notation for the accounting formulas below.

\noindent\textbf{Stage~1 (statement compilation).}
For each dataset item processed in Stage~1, the pipeline first generates a Lean statement skeleton (allowing proof placeholders) and then runs a verifier call on the affected file.
If verification fails, a local repair routine is invoked and the file is re-verified.
We cap the number of local repair rounds per item by a small constant $K$ (a fixed hyperparameter in the experiment configuration).
As a result, each Stage~1 item triggers at least one verifier call and at most $1+K$ verifier calls, with early termination when the file verifies.

\noindent\textbf{Stage~2 (proof completion).}
Stage~2 is organized as a bounded planning-and-execution loop.
A planning step proposes a structured proof strategy, and an execution step applies proof edits for the target hole; the file is then verified.
If execution fails in a way that suggests a flawed plan, the system may request replanning; if verification fails due to local compilation issues, the system may attempt bounded local repairs.
To make accounting explicit, we use two caps:
$C$ is the maximum number of planning rounds (including the initial plan), and $R$ is the maximum number of executor attempts per plan.
This yields the conservative upper bound
$
B \;=\; R \cdot C
$
on the number of executor-driven candidate proof patches for a single target.
This is an upper limit; in practice, targets often terminate early once solved or once a stopping condition is reached.

\noindent\textbf{Optional file splitting overhead.}
In some proof runs, a file may be split into multiple parts when it exceeds a line-count threshold, and additional verification checks may be triggered as part of that process.
When splitting is enabled, we count the splitter's model calls as oracle calls (see \S\ref{app:cost:oracle}) and we count any compilation checks performed during or after splitting as verifier calls.

\subsection{Reporting total verifier calls}
\label{app:cost:total_calls}

\noindent\textbf{Per-run totals from metrics events.}
Verifier calls are reconstructed from the metrics event log described in Appendix~\ref{app:json_schema:metrics}.
Each verifier invocation emits exactly one verifier-check event (in our logs this event is labeled \texttt{lean\_check}).
For a single run $r$, let $\mathsf{Events}(r)$ be the multiset of metrics records for that run, and define
\[
V_r \;=\; \#\{\, e \in \mathsf{Events}(r)\;:\; e.\texttt{event}=\texttt{lean\_check}\,\}.
\]
When a run is interrupted and resumed, each resumed segment produces its own run id; the corpus-level total is computed by summing $V_r$ across all run ids included in the reported experiment.

\noindent\textbf{Stage~1 metrics variant.}
Some Stage~1 runs in our archived artifacts were produced with a metrics schema that does not emit explicit \texttt{lean\_check} events.
For these runs, verifier-call totals can still be reconstructed because each processed item triggers one initial file verification, and each bounded repair attempt triggers one additional verification.
Let $\mathcal{I}(r)$ be the set of Stage~1 items completed in run $r$ (equivalently, the set of \texttt{item\_end} events), and let $\texttt{b\_attempts}(i)$ be the recorded number of bounded repair attempts for item $i$.
We then compute
\[
V_r \;=\; |\mathcal{I}(r)| \;+
\sum_{i\in\mathcal{I}(r)} \texttt{b\_attempts}(i).
\]
When \texttt{lean\_check} events are present (as in Stage~2 runs), we use the direct event-count definition above.

\noindent\textbf{Aggregating by corpus and stage.}
For each corpus/stage pair reported in the main paper, we aggregate (i) the number of targets processed, (ii) the number solved, and (iii) the total verifier calls.
For Stage~1, ``targets'' refers to the number of dataset items processed in the aggregated runs; ``solved'' may either be the number of items whose target file verified at item end, or simply equal to targets if the pipeline configuration guarantees a verified output per item.
For Stage~2, ``targets'' refers to the number of proof targets attempted under the reported setting, and ``solved'' refers to the number of targets whose designated proof hole was successfully closed under budget.

\noindent\textbf{Summary table.} Table~\ref{tab:app:total_verifier_calls} reports the aggregated target counts, solve counts, and total verifier calls for each corpus/stage setting used in the main paper.

\begin{table*}[t]
\caption{Total verifier calls and verifier-normalized cost.}
\label{tab:app:total_verifier_calls}
\centering
\small
\setlength{\tabcolsep}{3pt}
\renewcommand{\arraystretch}{1.05}
\resizebox{\textwidth}{!}{%
\begin{tabular}{@{}>{\raggedright\arraybackslash}m{0.22\textwidth}ccccc>{\raggedright\arraybackslash}m{0.24\textwidth}@{}}
\toprule
Corpus & Stage & Targets & Solved & Total verifier calls & Calls/solved & Notes \\
\midrule
\multirow[c]{2}{*}{\shortstack[l]{Textbooks:\\Real Analysis}} & 1 & 416 & 416 & 592 & 1.42 & \shortstack[l]{SCC $=100\%$\\\texttt{sorry} allowed} \\
& 2 & 339 & 339 & 628 & 1.85 & \shortstack[l]{Matched statements\\PSR $=100\%$} \\
\multirow[c]{2}{*}{\shortstack[l]{Textbooks:\\Convex Analysis}} & 1 & 560 & 560 & 605 & 1.08 & \shortstack[l]{SCC $=100\%$\\\texttt{sorry} allowed} \\
& 2 & 499 & 499 & 1065 & 2.13 & \shortstack[l]{Matched statements\\PSR $=100\%$} \\
\multirow[c]{2}{*}{Paper} & 1 & 67 & 67 & 81 & 1.20 & \shortstack[l]{SCC $=100\%$\\\texttt{sorry} allowed} \\
& 2 & 37 & 37 & 77 & 2.08 & \shortstack[l]{Matched statements\\PSR $=100\%$} \\
\shortstack[l]{Benchmark: FATE-H\\(auto)} & 2 & 100 & 96 & 283 & 2.95 & \shortstack[l]{One instance is incorrect\\see \S\ref{sec:exp:failure}} \\
\shortstack[l]{Benchmark: FATE-H\\(+ lemma map)} & 2 & 100 & 97 & 368 & 3.79 & \shortstack[l]{31 decl hints\\see Table~\ref{tab:fateh_psr}} \\
\bottomrule
\end{tabular}
}
\end{table*}

\subsection{Paper metrics: definitions and log reconstruction}
\label{app:cost:metrics_defs}

This subsection defines the primary paper metrics and explains how they are reconstructed from the structured logs in Appendix~\ref{app:json_schema}.
Throughout, we treat the Lean toolchain under the pinned environment $\mathcal{E}$ as the only acceptance signal: a proposed change ``counts'' only if it is committed by the pipeline and the subsequent verifier check reports zero error-level diagnostics.

\noindent\textbf{PB (project buildability).}
PB is true iff the final generated project builds under $\mathcal{E}$ (allowing \texttt{sorry} in Stage~1).
Operationally, PB is assessed on the end-to-end project state produced by the pipeline, not on an individual item or file.
In our instrumentation, PB is recorded as a run-level outcome and can also be reproduced by running a project build in the pinned environment.

\noindent\textbf{SCC (statement compile coverage).}
Let $\mathcal{B}$ be the set of Stage~1 atomic blocks processed in a corpus run.
A block is counted as compiled iff its enclosing Lean file elaborates under $\mathcal{E}$ after the block insertion/repair loop terminates.
We report $\mathrm{SCC}=|\mathcal{B}_{ok}|/|\mathcal{B}|$.
In logs, $\mathcal{B}$ is reconstructed from the processed item inventory; success is reconstructed from item-end status markers and the final verifier outcome for the affected file.

\noindent\textbf{ARR (average repair rounds per block).}
For each block $b\in\mathcal{B}$, let $a(b)$ be the number of candidate repair patches attempted after the initial insertion (blocks that elaborate immediately contribute $a(b)=0$).
We report $\mathrm{ARR}=\frac{1}{|\mathcal{B}_{ok}|}\sum_{b\in\mathcal{B}_{ok}} a(b)$, i.e., the mean repair-attempt count over successfully processed blocks.
In logs, $a(b)$ can be reconstructed either by counting repair-result events attributable to that block or by counting the verifier checks that occur inside the bounded verify$\rightarrow$repair loop for the block (minus the initial post-insertion check).

\noindent\textbf{PSR (proof success rate under matched statements).}
Let $\mathcal{H}$ be the evaluated set of proof obligations under matched statements (the statement layer is frozen and not edited during proof search).
A hole counts as closed iff (i) the designated \texttt{sorry} is eliminated in the target file and (ii) the verifier still reports zero error-level diagnostics for that file under $\mathcal{E}$.
We define $\mathrm{PSR}=|\mathcal{H}_{closed}|/|\mathcal{H}|$.
In logs, $\mathcal{H}$ is reconstructed from the evaluated task inventory together with per-target identifiers, and closure is reconstructed from the first verifier check that both verifies and decreases the hole count for that target.

\noindent\textbf{Auxiliary reporting on FATE-H.}
Because one FATE-H instance is a benchmark issue whose Lean statement is already incorrect (and thus not provable as stated), we sometimes additionally report success excluding this instance.
Concretely, the fully automatic setting solves $96/100$ instances (Table~\ref{tab:fateh_psr}), while excluding the incorrect instance yields $96/99\approx 97.0\%$ solved; see \S\ref{sec:exp:failure}.

\subsection{FATE-H per-problem length and solved status}
\label{app:cost:fateh_length}

\begin{figure*}[t]
  \centering
  \includegraphics[width=0.95\linewidth]{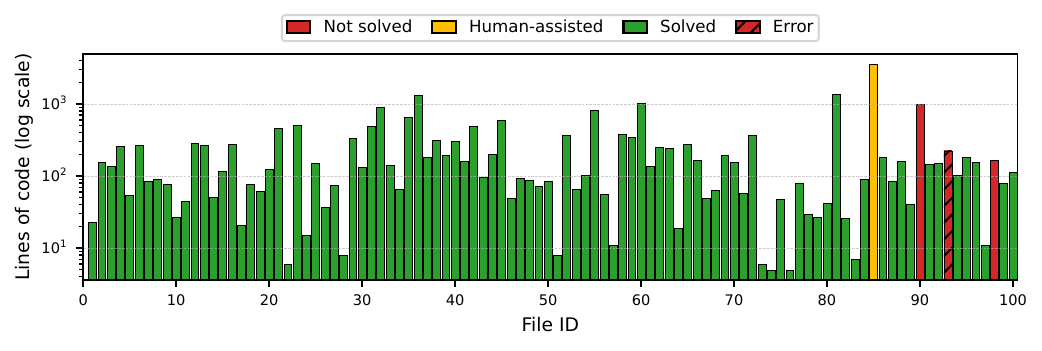}
  \caption{FATE-H per-problem code length (non-empty lines) and outcome category. Colors indicate outcome: green = solved automatically, yellow = solved with lemma-map supervision, red = unsolved; the single wrong-statement instance is shown with a distinct style (see \S\ref{sec:exp:failure}).}
  \label{fig:app:fateh_length_solved}
\end{figure*}

To contextualize aggregate success rates on FATE-H, Figure~\ref{fig:app:fateh_length_solved} reports, for each of the 100 problems, a simple code-length proxy together with the outcome category under our Stage~2 runs.
We define the length of a problem as the number of \emph{non-empty lines of Lean code} in its source file, measured after preprocessing and proof blanking but before any successful proof patch is applied.
Because FATE-H contains many problems, we plot a bar chart: each bar corresponds to one problem (x-axis), and bar length is the line count (y-axis).

\noindent\textbf{Outcome categories.}
Bars are colored by outcome: \emph{solved fully automatically} (green), \emph{solved with light expert help via the lemma map} (yellow), and \emph{unsolved} (red).
One additional category marks the single \emph{wrong-statement} instance (``Error''), discussed in the Failure Analysis (\S\ref{sec:exp:failure}); this instance is not solvable under $\mathcal{E}$ as stated and is visualized with a distinct styling (e.g., a separate color or hatch pattern).
In our reported runs, this yields 96 solved automatically, 1 solved with the lemma map, 2 remaining open, and 1 wrong instance (Table~\ref{tab:fateh_psr}).

\subsection{Non-verifier oracle calls and their frequency}
\label{app:cost:oracle}

In addition to verifier calls, the system makes non-verifier \emph{oracle calls} to the model through Codex CLI.
These calls produce candidate Lean edits (statement skeletons, proof patches, repairs) or structured plans that guide subsequent edits.
Oracle calls are not counted as verifier calls, but they can dominate wall-clock time and token usage, so we report their frequency and use them in robustness checks.

\noindent\textbf{What counts as an oracle call.}
An oracle call is any model invocation whose output can directly change the Lean project or the plan used to change it.
In Stage~1, oracle calls include the initial statement-synthesis call and any bounded repair calls used to make the file compile.
In Stage~2, oracle calls include planning calls, executor calls that apply proof edits, repair calls that address compilation failures after edits, and (when enabled) splitter calls that refactor large files into multiple parts.
In the final cleanup stage (when used), oracle calls include compilation-fix and warning-cleanup calls as well as plan/execution calls that eliminate remaining \texttt{sorry} tokens.

\noindent\textbf{How oracle calls are counted in logs.}
Oracle calls are reconstructed from the metrics event log (Appendix~\ref{app:json_schema:metrics}).
Each model invocation emits one oracle-result event whose label identifies the agent role (planner/executor/repair/splitter).
We count oracle calls by counting these oracle-result events within each run.
If a reported corpus/stage aggregates multiple run ids (e.g., due to resumption), we sum oracle-call counts across those run ids, matching the aggregation used for verifier calls.

\noindent\textbf{Stage~1 metrics variant.}
Some Stage~1 runs in our archived artifacts were produced with a metrics schema that does not emit explicit repair-call events for the bounded repair agent.
However, the Stage~1 control flow makes oracle-call totals reconstructible from the same per-item repair counter: each processed item always triggers one statement-synthesis model call, and each bounded repair attempt corresponds to one additional repair model call.
Let $\mathcal{I}(r)$ and \texttt{b\_attempts} be as above; we compute
\[
Q_r \;=\; |\mathcal{I}(r)| \;+
\sum_{i\in\mathcal{I}(r)} \texttt{b\_attempts}(i),
\]
which matches the verifier-call reconstruction for Stage~1.

\noindent\textbf{Note on Lean LSP/MCP tool queries.}
Some runs issue lightweight Lean IDE queries during editing (goal inspection, hover/type information, local search, completions, and small snippet execution).
These are not model invocations and are not emitted as metrics events, so they are excluded from the oracle-call totals above.
They may contribute to wall-clock time and local compute, but they are outside the verifier-normalized and token-based accounting reported in this appendix.

\noindent\textbf{Summary table.} Table~\ref{tab:app:oracle_call_freq} reports oracle-call frequency alongside verifier calls, and normalizes by the number of targets to make stages and corpora comparable.

\begin{table*}[t]
\caption{Oracle-call frequency.}
\label{tab:app:oracle_call_freq}
\centering
\small
\setlength{\tabcolsep}{3pt}
\renewcommand{\arraystretch}{1.08}
\resizebox{\textwidth}{!}{%
\begin{tabular}{@{}>{\raggedright\arraybackslash}p{0.22\textwidth}ccccc>{\raggedright\arraybackslash}p{0.24\textwidth}@{}}
\toprule
Corpus & Stage & Verifier calls & Oracle calls & Calls/target & Notes \\
\midrule
\multirow{2}{*}{\shortstack[l]{Textbooks:\\Real Analysis}}  & 1 & 592 & 592  & 1.42 & Statement synthesis + bounded repairs \\
& 2 & 628 & 1263 & 3.73 & Planner/executor/repair calls \\
\multirow{2}{*}{\shortstack[l]{Textbooks:\\Convex Analysis}} & 1 & 605 & 605  & 1.08 & Statement synthesis + bounded repairs \\
& 2 & 1065 & 2001 & 4.01 & Planner/executor/repair (+ optional split) \\
\multirow{2}{*}{Paper}                      & 1 & 81  & 81   & 1.21 & Statement synthesis + bounded repairs \\
& 2 & 77  & 148  & 4.00 & Planner/executor/repair calls \\
\shortstack[l]{Benchmark: FATE-H\\(auto)}                    & 2 & 283 & 339  & 3.39 & Proof-only; statements already elaborate \\
\shortstack[l]{Benchmark: FATE-H\\(+ lemma map)}             & 2 & 368 & 479  & 4.79 & Proof-only; conditioned on decl hints \\
\bottomrule
\end{tabular}
}
\end{table*}

\subsection{Token-based cost for oracle calls}
\label{app:cost:tokens}

Oracle-call frequency measures how many model invocations are made, but it does not capture their size.
We therefore additionally report token usage as a proxy for model-side cost.

\noindent\textbf{Token totals.}
When per-call token counts are recorded directly in metrics events, total tokens can be computed by summing those per-call fields over oracle-result events in the run.
To make accounting robust to logging variants, we treat the token-backfill artifact described in Appendix~\ref{app:json_schema:token_backfill} as the canonical source of token totals when available, because it reconstructs per-task token usage directly from stable token markers in the per-call agent logs.

\noindent\textbf{Summary table.} Table~\ref{tab:app:token_usage} reports total token usage for oracle calls as a complementary proxy for model-side cost, together with a per-solved normalization for the Stage~2 settings.

\begin{table*}[ht]
\caption{Token usage for oracle calls.}
\label{tab:app:token_usage}
\centering
\small
\setlength{\tabcolsep}{3pt}
\renewcommand{\arraystretch}{1.05}
\begin{tabular}{@{}m{0.27\textwidth}cccc@{}}
\toprule
Corpus & Stage & Oracle calls & Total tokens & Tokens/solved \\
\midrule
\multirow{2}{*}{Textbooks: Real Analysis}  & 1 & 592  & 20429061  & 49108  \\
                                          & 2 & 1263 & 140040564 & 413099 \\
\multirow{2}{*}{Textbooks: Convex Analysis} & 1 & 605  & 28421469  & 50753  \\
                                           & 2 & 2001 & 307966089 & 617167 \\
\multirow{2}{*}{Paper}                      & 1 & 81   & 2875173   & 42913  \\
                                           & 2 & 148  & 16362141  & 442220 \\
\shortstack[l]{Benchmark: FATE-H\\(auto)} & 2 & 339 & 44177658 & 460184 \\
\shortstack[l]{Benchmark: FATE-H\\(+ lemma map)} & 2 & 479 & 53766276 & 554292 \\
\bottomrule
\end{tabular}
\end{table*}

\subsection{Fractional accounting robustness check}
\label{app:cost:fractional}

Verifier calls are our primary compute unit, but model invocations can also contribute nontrivial cost.
To check that our conclusions are not sensitive to how oracle calls are priced, we report a conservative \emph{fractional accounting} view that charges each oracle call as a fixed fraction of a verifier call.

\noindent\textbf{Definition.}
For a run (or an aggregation of runs), let $V$ be the total number of verifier calls and let $Q$ be the total number of oracle calls (as defined above).
For a chosen fraction $\alpha\in[0,1]$, we define a verifier-equivalent compute total
\[
\mathrm{Cost}_{\alpha} \;=\; V \;+\; \alpha \cdot Q.
\]
We report $\mathrm{Cost}_{\alpha}$ (and optionally $\mathrm{Cost}_{\alpha}$/solved) for a small set of $\alpha$ values that cover ``oracle is cheap'' to ``oracle is comparable'' regimes.

\noindent\textbf{Reporting.}
Unless otherwise stated, we use $\alpha \in \{0.05,\,0.10,\,0.25\}$.
The corresponding totals are computed from the same reconstructed $V$ and $Q$ used in Tables~\ref{tab:app:total_verifier_calls} and \ref{tab:app:oracle_call_freq}.

\noindent\textbf{Summary table.} Table~\ref{tab:app:fractional_cost} reports verifier-equivalent compute totals under fractional accounting for representative choices of $\alpha$, using the same $V$ and $Q$ aggregates reported above.

\begin{table}[ht]
\caption{Fractional accounting: verifier-equivalent compute $\mathrm{Cost}_{\alpha}=V+
\alpha Q$}
\label{tab:app:fractional_cost}
\centering
\small
\setlength{\tabcolsep}{4pt}
\renewcommand{\arraystretch}{1.05}
\begin{tabular}{@{}p{0.38\linewidth}rrrr@{}}
\toprule
Corpus/setting & $V$ & $Q$ & $\mathrm{Cost}_{0.10}$ & $\mathrm{Cost}_{0.25}$ \\
\midrule
Textbooks (Real Analysis, Stage~2) & 628 & 1263 & 754.30 & 943.75 \\
Textbooks (Convex Analysis, Stage~2) & 1065 & 2001 & 1265.10 & 1565.25 \\
Paper (Stage~2) & 77 & 148 & 91.80 & 114.00 \\
FATE-H (auto, Stage~2) & 283 & 339 & 316.90 & 367.75 \\
\bottomrule
\end{tabular}
\end{table}

\section{Qualitative Showcase: Navigable Artifacts with Provenance and Dependency Governance}
\label{app:showcase}

This appendix provides a qualitative view of what the end-to-end system produces \emph{as a library artifact}, complementing the quantitative results in the main text.
While Table~\ref{tab:proj_stats} already reports the primary scale statistics (files/decls/LoC and end-to-end buildability), we focus here on three properties that are hard to convey with a single success rate: (i) a closed-loop workflow with explicit governance points, (ii) ToC-faithful \emph{navigability} from a source section to a small, checkable edit scope in Lean, and (iii) provenance- and dependency-centric indicators that support auditing and maintenance.

\setlength{\textfloatsep}{8pt plus 2pt minus 2pt}
\setlength{\floatsep}{6pt plus 2pt minus 2pt}
\setlength{\intextsep}{6pt plus 2pt minus 2pt}
\setlength{\dbltextfloatsep}{8pt plus 2pt minus 2pt}
\setlength{\dblfloatsep}{6pt plus 2pt minus 2pt}
\captionsetup{skip=4pt}

\subsection{Workflow overview: a closed loop, not one-shot generation}
\label{app:showcase:pipeline}

\begin{figure*}[t]
\centering
\includegraphics[width=\textwidth]{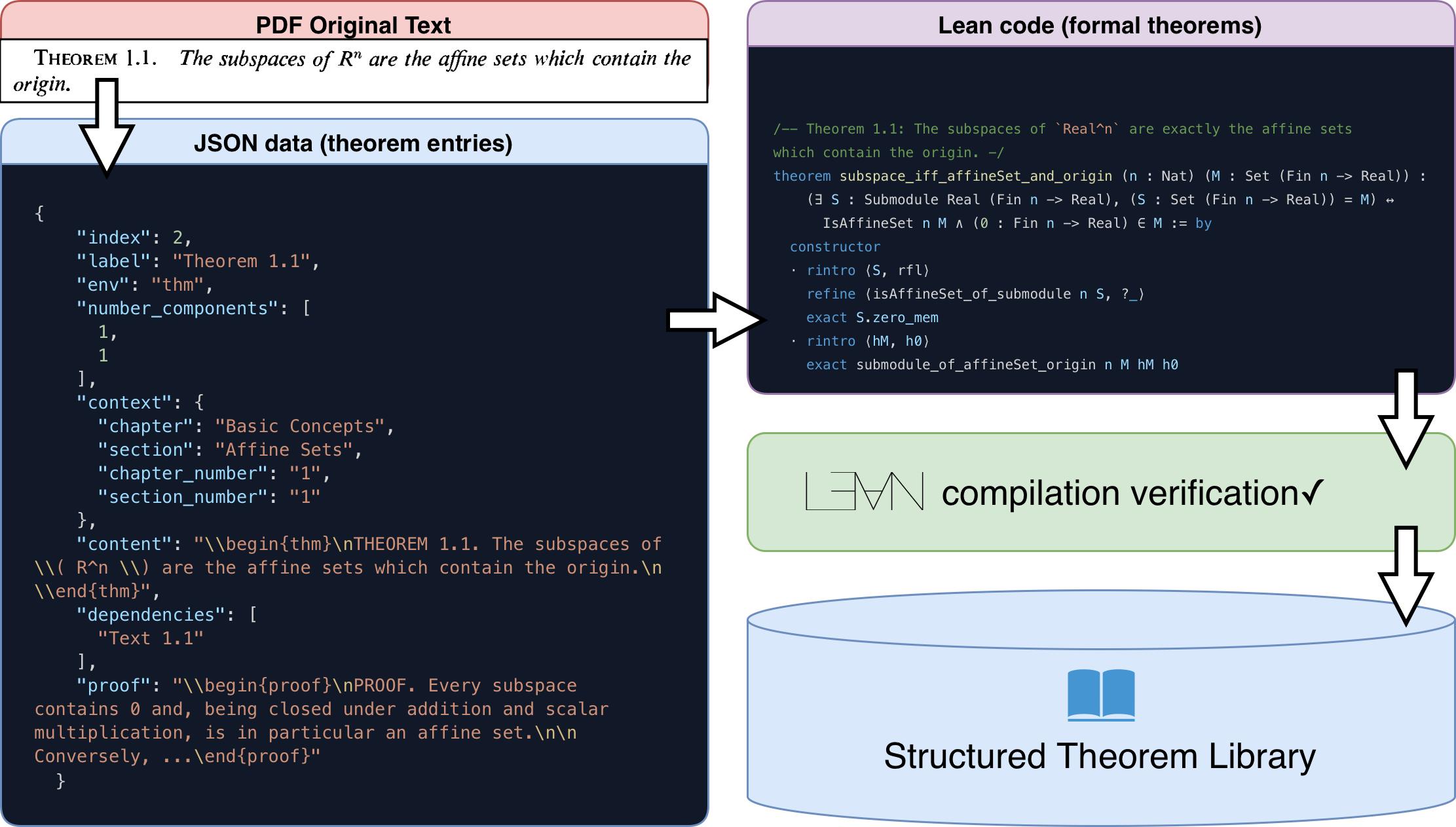}%
\caption{\textbf{System capability manifesto (workflow).}
A verifier-in-the-loop pipeline that turns PDF-derived structure into a buildable Lean project with (i) provenance anchoring for statement auditability, (ii) human gatekeeping at governance points, and (iii) accept/revert refinement driven solely by Lean diagnostics, yielding a queryable repository of verified declarations rather than isolated one-off proofs.}
\label{fig:app:workflow_overview}
\end{figure*}

The system treats project-scale formalization as a \emph{closed loop} driven by the Lean toolchain: edits are accepted only when the verifier certifies a strict improvement in a build- and hole-based objective.
This design gives two practical guarantees that matter at scale.
First, failures are localized (file/section/part granularity), enabling fast revert-and-retry without corrupting the rest of the project.
Second, outputs are not merely a pile of proofs, but a structured repository with explicit provenance anchors and a controlled import footprint, so that downstream users can audit and maintain the artifact. Figure~\ref{fig:app:workflow_overview} provides a compact, governance-oriented view of this loop, complementing the end-to-end pipeline diagram in Figure~\ref{fig:m2f_pipeline}.

\subsection{Source-to-project alignment at section granularity}
\label{app:showcase:alignment}

A central ``project-scale'' requirement is \emph{navigability}: readers should be able to move from a source section to the corresponding Lean files with minimal cognitive overhead.
For Rockafellar's \emph{Convex Analysis}, we mirror the source section numbering (\S1--\S15) at the module level and display the generated Lean file tree with indentation (not full paths).
When a section is too large to edit and verify reliably, the workflow performs a structure-preserving split into \texttt{\_partK} modules, while keeping a stable section-level entry module as the public import target.

\begin{figure*}[t]
\centering
\includegraphics[width=\textwidth]{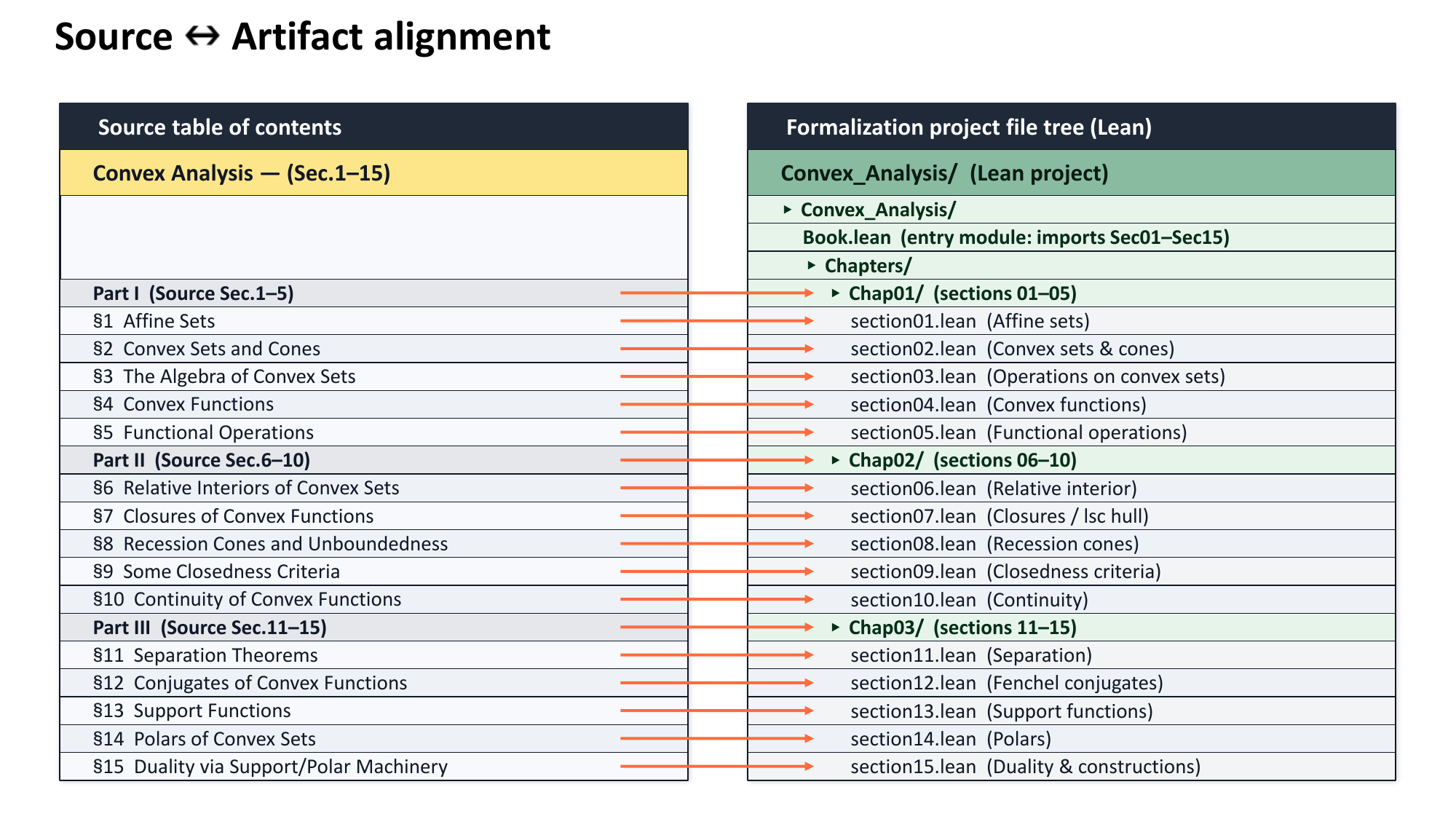}
\caption{\textbf{System capability manifesto (navigability and locality).}
Rockafellar \emph{Convex Analysis} (\S1--\S15) is re-indexed into a ToC-faithful Lean project.
The right panel is an indented file tree (not full paths) illustrating structure-preserving splitting (e.g., a single source section expanded into many \texttt{\_partK} modules) while keeping a stable section module for downstream imports and verifier-local edits.}
\label{fig:app:ca_alignment}
\end{figure*}

\subsection{Workflow capability indicators beyond raw scale}
\label{app:showcase:indicators}

The main text already reports corpus scale and buildability (Table~\ref{tab:proj_stats}).
Here we focus on \emph{governance} indicators that are more directly tied to whether the workflow remains controllable as projects grow: how aggressively the system can split a single section without breaking the public module interface, how large the import surface becomes, and whether the artifact carries enough provenance anchors to support auditing.
The counts below are computed from our artifact snapshot; where appropriate we report coarse magnitudes (e.g., ``160+'').

\begin{table}[t]
\caption{\textbf{System capability manifesto (governance at scale).}
Indicators that emphasize (i) edit locality via structure-preserving splitting, (ii) dependency governance via an inspectable import footprint, and (iii) auditability via provenance anchors attached to declarations. Scale (files/decls/LoC, PB/PSR) is reported in Table~\ref{tab:proj_stats}.}
\label{tab:app:workflow_indicators}
\centering
\small
\setlength{\tabcolsep}{3pt}
\renewcommand{\arraystretch}{1.08}
\begin{tabular}{@{}p{0.40\linewidth}p{0.14\linewidth}p{0.24\linewidth}p{0.14\linewidth}@{}}
\toprule
Indicator & Real Analysis & Convex Analysis (\S1--\S15) & Paper \\
\midrule
Max structure-preserving split factor & 6-way & 23-way & 10-way \\
Unique imported modules (approx.)     & 50      & 160+    & 25+ \\
Provenance anchors in docstrings (approx.) & 400+ & 600+ & 70+ \\
Remaining \texttt{sorry} (closure)    & 0       & 0       & 0 \\
\bottomrule
\end{tabular}
\end{table}

\noindent\textbf{Highlights.}
In the \emph{Convex Analysis} artifact, the workflow cleanly executes a \textbf{23-way} structure-preserving split of a single section, attaches \textbf{600+} provenance anchors, and manages \textbf{160+} unique imports---a concrete demonstration of dependency governance and auditability at textbook scale. These indicators reflect the system's ability to keep edit scopes small and dependencies inspectable while still producing end-to-end verified projects.

\subsection{Representative verified code excerpts}
\label{app:showcase:snippets}

We include three excerpts that illustrate complementary sources of complexity in project-scale formalization.
First, we show a textbook-level interface definition from \emph{Convex Analysis} that is imported across multiple chapters, illustrating how the artifact exposes reusable notions behind a small, stable API surface (here, the definition of a supporting hyperplane).
Second, we include a classical result from \emph{Real Analysis} (the mean value theorem) as a representative example of a long proof whose correctness must remain stable under localized edits and refactoring; for space, we elide internal steps with ellipses.
Third, we present a paper-style theorem with heavy parameterization and layered hypotheses (Algorithm~3.11 / Theorem~1.4.1), which stresses scoping, abbreviation management, and long-range dependencies across the library.
For robustness under \texttt{pdfLaTeX}, the listings are lightly normalized (e.g., we use ASCII fallbacks or omit nonessential proof details where appropriate); this affects only presentation and not the underlying Lean development.

\lstdefinestyle{LeanASCII}{
  basicstyle=\ttfamily\footnotesize,
  columns=fullflexible,
  keepspaces=true,
  breaklines=true,
  breakatwhitespace=true,
  showstringspaces=false,
  literate=
    {→}{{$\rightarrow$}}1
    {ℝ}{{$\mathbb{R}$}}1
    {ℕ}{{$\mathbb{N}$}}1
    {∀}{{$\forall$}}1
    {∃}{{$\exists$}}1
    {∈}{{$\in$}}1
    {∑}{{$\sum$}}1
    {∧}{{$\land$}}1
    {≠}{{$\neq$}}1
    {≤}{{$\le$}}1
    {≥}{{$\ge$}}1
    {σ}{{$\sigma$}}1
    {μ}{{$\mu$}}1
    {φ}{{$\phi$}}1
    {ε}{{$\epsilon$}}1
    {‖}{{$\Vert$}}1
    {•}{{$\bullet$}}1
    {₁}{{$_1$}}1
    {ₗ}{{$_\ell$}}1
    {⬝}{{$\cdot$}}1
    {ᵥ}{{$_v$}}1
}
\begin{tcolorbox}[
  colback=gray!6,
  colframe=black!60,
  boxrule=0.4pt,
  arc=1mm,
  left=1.5mm, right=1.5mm, top=0.8mm, bottom=0.8mm,
  before skip=6pt, after skip=6pt
]
\begin{lstlisting}[style=LeanASCII]
/- Convex Analysis (textbook): multi-file dependency surface + interface lemma. -/
import Mathlib
import FormalBook.Chapters.Chap01.section01_part1
import FormalBook.Chapters.Chap02.section09_part3
import FormalBook.Chapters.Chap03.section11_part4

/-- Text 11.3.2: A supporting hyperplane to `C` is a hyperplane which is the boundary of a
supporting half-space to `C`. Equivalently, a supporting hyperplane has the form
`H = {x | x ⬝ᵥ b = beta}` with `b ≠ 0`, such that `x ⬝ᵥ b ≤ beta` for every `x ∈ C` and
`x ⬝ᵥ b = beta` for at least one `x ∈ C`. -/
def IsSupportingHyperplane (n : Nat) (C H : Set (Fin n -> Real)) : Prop :=
  ∃ (b : Fin n → Real) (beta : Real),
    b ≠ 0 ∧
      H = {x : Fin n → Real | x ⬝ᵥ b = beta} ∧
        (∀ x, x ∈ C → x ⬝ᵥ b ≤ beta) ∧ ∃ x, x ∈ C ∧ x ⬝ᵥ b = beta
\end{lstlisting}
\end{tcolorbox}

\begin{tcolorbox}[
  colback=gray!6,
  colframe=black!60,
  boxrule=0.4pt,
  arc=1mm,
  left=1.5mm, right=1.5mm, top=0.8mm, bottom=0.8mm,
  before skip=6pt, after skip=6pt
]
\begin{lstlisting}[style=LeanASCII]
/-- Theorem 4.2.4 (Mean value theorem): If `f : [a, b] → ℝ` is continuous on
the closed interval and differentiable on the open interval, then some
`c ∈ (a, b)` satisfies `f b - f a = deriv f c * (b - a)`. -/
theorem mean_value_theorem
    {f : ℝ → ℝ} {a b : ℝ} (h₁ : a < b)
    (hcont : ContinuousOn f (Set.Icc a b))
    (hdiff : DifferentiableOn ℝ f (Set.Ioo a b)) :
    ∃ c ∈ Set.Ioo a b, f b - f a = deriv f c * (b - a) := by
  ...
\end{lstlisting}
\end{tcolorbox}

\begin{tcolorbox}[
  colback=gray!6,
  colframe=black!60,
  boxrule=0.4pt,
  arc=1mm,
  left=1.5mm, right=1.5mm, top=0.8mm, bottom=0.8mm,
  before skip=6pt, after skip=6pt
]
\begin{lstlisting}[style=LeanASCII]
/-- Theorem 1.4.1.
Assume the structural model (2.2) and that `fhat` is differentiable with M-Lipschitz gradient on
`Q1`. Let `d1` be a prox-function of `Q1` with parameter `σ1 > 0` and prox-diameter `D1` as in
(4.3_D1). Let `d2` be a prox-function of `Q2` with strong convexity parameter `σ2 > 0` and define
`D2 := max_{u ∈ Q2} d2 u` as in Proposition 2.7. Apply Algorithm 3.11 to the smoothed problem
(4.1) with `μ = μ(N) = (2‖A‖_{1,2}/(N+1)) * sqrt(D1/(σ1 σ2 D2))` (equation (thm3_muN)). After `N`
iterations define `\hat x := y_N` and
`\hat u := ∑_{i=0}^N 2(i+1)/((N+1)(N+2)) u_μ(x_i)` (4.2). Then `0 ≤ f(\hat x) - φ(\hat u)` and the
duality-gap bound (4.3) holds, and consequently the ε-solution bound (4.4) holds. -/
theorem algorithm311_duality_gap_bound {E1 E2 : Type*} [NormedAddCommGroup E1]
    [NormedSpace ℝ E1] [FiniteDimensional ℝ E1] [NormedAddCommGroup E2] [NormedSpace ℝ E2]
    [FiniteDimensional ℝ E2] (Q1 : Set E1) (Q2 : Set E2)
    (A : E1 →L[ℝ] (E2 →L[ℝ] ℝ)) (fhat : E1 → ℝ) (phihat : E2 → ℝ)
    (d1 : E1 → ℝ) (d2 : E2 → ℝ) (σ1 σ2 M D1 : ℝ) (xSeq ySeq : ℕ → Q1) (uμ : E1 → E2)
    (N : ℕ) (hatu : Q2) : let A' : E1 →ₗ[ℝ] Module.Dual ℝ E2 :=
      { toFun := fun x => (A x).toLinearMap
        map_add' := by
          intro x y;ext u;simp
        map_smul' := by
          intro c x;ext u;simp }
    let D2 : ℝ := sSup (d2 '' Q2)
    let μ : ℝ := (2 * OperatorNormDef A' / ((N : ℝ) + 1)) *
        Real.sqrt (D1 / (σ1 * σ2 * D2))
    let fbar : E1 → ℝ := SmoothedObjective Q2 A phihat d2 μ fhat
    let f : E1 → ℝ := fun x => fhat x + sSup ((fun u => A x u - phihat u) '' Q2)
    let φ : Q2 → ℝ := AdjointFormPotential Q1 Q2 A fhat phihat
    LipschitzOnWith (Real.toNNReal M) (fun x => fderiv ℝ fhat x) Q1 →
    0 < σ1 → 0 < σ2 → StrongConvexOn Q1 σ1 d1 → StrongConvexOn Q2 σ2 d2 →
    IsProxDiameterBound Q1 d1 D1 → OptimalSchemeAlgorithm Q1 fbar d1
      (M + (1 / (μ * σ2)) * (OperatorNormDef A') ^ 2) σ1 xSeq ySeq →
    (∀ x, IsSmoothedMaximizer Q2 A phihat d2 μ x (uμ x)) →
    (hatu : E2) = Finset.sum (Finset.range (N + 1)) (fun i =>
        (2 * ((i : ℝ) + 1) / (((N : ℝ) + 1) * ((N : ℝ) + 2))) •
          uμ (xSeq i : E1)) →
    ConvexOn ℝ Q1 fhat → DifferentiableOn ℝ fhat Q1 →
    ConvexOn ℝ Q2 phihat →
    (∀ x,
      fderiv ℝ (SmoothedMaxFunction Q2 A phihat d2 μ) x =
        (AdjointOperator A' (uμ x)).toContinuousLinearMap) →
    0 ≤ M → 0 ≤ D1 →
    0 < M + (1 / (μ * σ2)) * (OperatorNormDef A') ^ 2 →
    IsClosed Q1 → IsOpen Q1 → ConvexOn ℝ Q1 fbar → DifferentiableOn ℝ fbar Q1 →
    0 ≤ μ → (∀ u ∈ Q2, 0 ≤ d2 u) →
    (∀ x, BddAbove ((fun u => A x u - phihat u) '' Q2)) →
    BddAbove (d2 '' Q2) → Q2.Nonempty →
    (∀ u, BddBelow ((fun x => A x u + fhat x) '' Q1)) →
    0 ≤ f (ySeq N : E1) - φ hatu ∧
      f (ySeq N : E1) - φ hatu ≤
        (4 * OperatorNormDef A' / ((N : ℝ) + 1)) * Real.sqrt (D1 * D2 / (σ1 * σ2)) +
          (4 * M * D1) / (σ1 * ((N : ℝ) + 1) ^ (2 : ℕ)) ∧
      ∀ ε > 0, (N : ℝ) ≥
          (4 * OperatorNormDef A' / ε) * Real.sqrt (D1 * D2 / (σ1 * σ2)) +
            2 * Real.sqrt (M * D1 / (σ1 * ε)) → f (ySeq N : E1) - φ hatu ≤ ε := by
    ...
\end{lstlisting}
\end{tcolorbox}